%% file: neurips_2023.tex
\documentclass{article}

\usepackage[final, nonatbib]{neurips_2023}

\input{shortcuts}

\title{SyncDiffusion: Coherent Montage via Synchronized Joint Diffusions}

\author{Yuseung Lee $\quad$
Kunho Kim $\quad$
Hyunjin Kim $\quad$
Minhyuk Sung \\[0.2em]
KAIST \\
{\tt\small \{phillip0701,kaist984,rlaguswls98,mhsung\}@kaist.ac.kr}
}

\begin{document}
\maketitle

\input{sections/00_abstract}

\input{sections/01_intro}

\input{sections/02_related}

\input{sections/03_method}
\input{sections/04_results}
\input{sections/05_conclusion}

{\small
\bibliographystyle{plain}
\bibliography{main}
}

\newif\ifpaper
\papertrue

\clearpage
\newpage

\section*{Appendix}
\input{sections/06_supp}

\end{document}

%% file: shortcuts.tex
\usepackage[linesnumbered,ruled,vlined]{algorithm2e}
\usepackage{amsfonts}       %
\usepackage{amsmath}
\usepackage{booktabs}       %
\usepackage{enumitem}
\usepackage[T1]{fontenc}    %
\usepackage{graphicx}
\usepackage{hyperref}       %
\usepackage[utf8]{inputenc} %
\usepackage{makecell}
\usepackage{microtype}      %
\usepackage{multirow}
\usepackage{nicefrac}       %
\usepackage{tabularx}
\usepackage{url}            %
\usepackage{xcolor}         %

\usepackage{longtable}
\usepackage{filecontents,catchfile,pgffor}

\usepackage{environ}

\newcommand*{\SyncDiffusion}[0]{\textsc{SyncDiffusion}}

\newcolumntype{Y}{>{\centering\arraybackslash}X}

\newcommand*{\V}[1]{\mathbf{#1}}
\newcommand*{\C}[1]{\mathcal{#1}}

\SetAlCapHSkip{0pt}
\SetKwInput{KwInput}{Inputs}
\SetKwInput{KwOutput}{Outputs}
\SetKwInOut{KwParam}{Parameters}

\definecolor{darkred}{rgb}{0.7,0.2,0.1}
\definecolor{darkgreen}{rgb}{0,0.7,0}
\definecolor{orange}{RGB}{255,127,0}
\definecolor{ourpurple}{RGB}{127,127,204}
\definecolor{palgreen}{RGB}{51,179,179}
\definecolor{magenta}{RGB}{199,21,133}

%% file: sections/00_abstract.tex
\vspace{-\baselineskip}
\begin{abstract}
\vspace{-0.5\baselineskip}
The remarkable capabilities of pretrained image diffusion models have been utilized not only for generating fixed-size images but also for creating panoramas. However, naive stitching of multiple images often results in visible seams. Recent techniques have attempted to address this issue by performing joint diffusions in multiple windows and averaging latent features in overlapping regions. However, these approaches, which focus on seamless montage generation, often yield incoherent outputs by blending different scenes within a single image. To overcome this limitation, we propose~\SyncDiffusion{}, a plug-and-play module that synchronizes multiple diffusions through gradient descent from a perceptual similarity loss. Specifically, we compute the gradient of the perceptual loss using the predicted denoised images at each denoising step, providing meaningful guidance for achieving coherent montages. Our experimental results demonstrate that our method produces significantly more coherent outputs for text-guided panorama generation compared to previous methods (66.35\% vs. 33.65\% in our user study) while still maintaining fidelity (as assessed by GIQA) and compatibility with the input prompt (as measured by CLIP score). 
We further demonstrate the versatility of our method across three plug-and-play applications: layout-guided image generation, conditional image generation and 360-degree panorama generation.
Our project page is at \textcolor{magenta}{\url{https://syncdiffusion.github.io}}.
\end{abstract}

\begin{figure}[h]
\centering
\vspace{-\baselineskip}
{
    \setlength{\fboxrule}{2pt}
    \setlength{\fboxsep}{0pt}
    \includegraphics[width=0.85\textwidth]{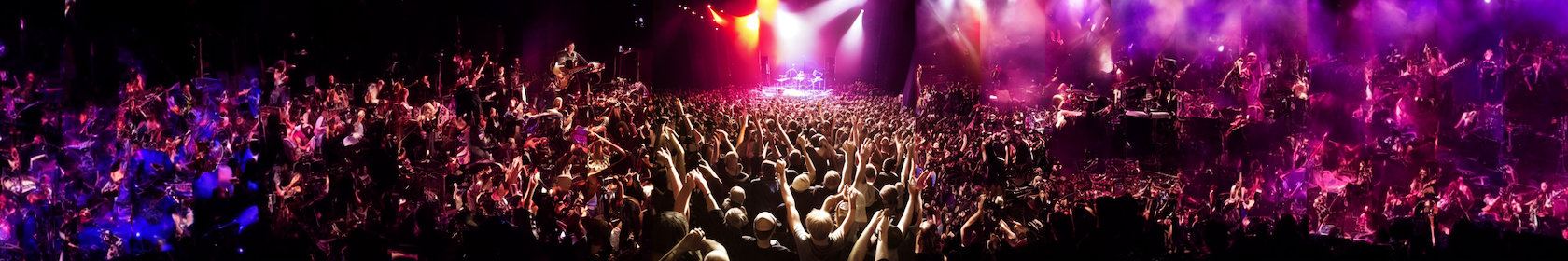}
    \includegraphics[width=0.85\textwidth]{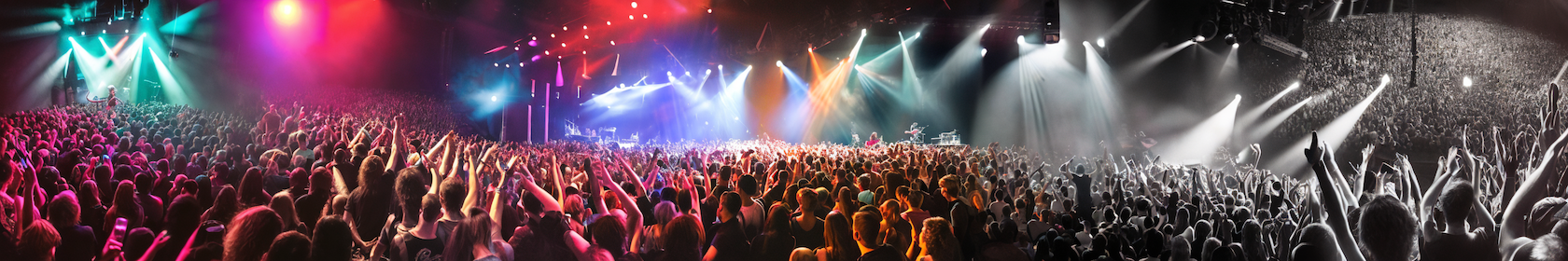}
    \fcolorbox{black}{black}{\includegraphics[width=0.85\textwidth]{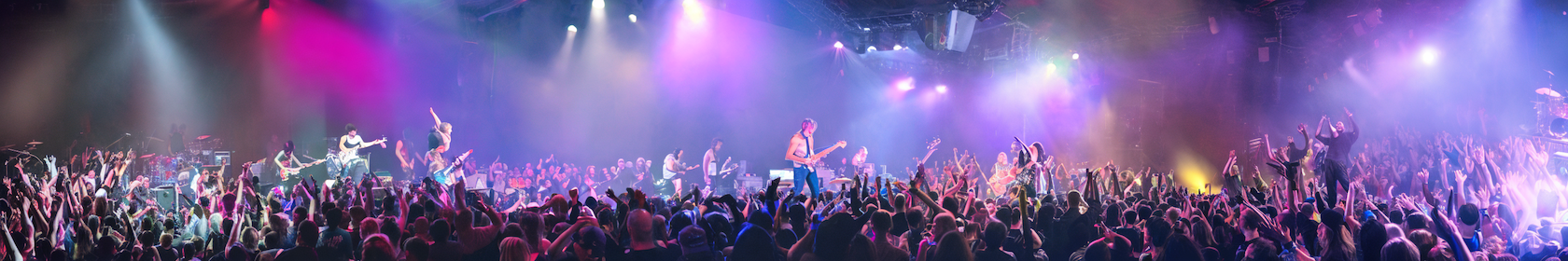}} 
}
\caption{
Comparison of panoramas generated with prompt \textit{``A photo of a rock concert''} by Blended Latent Diffusion~\cite{avrahami2023blended_latent} (top), MultiDiffusion~\cite{bartal2023multidiffusion} (middle), and our~\SyncDiffusion{} (bottom).
Blended Latent Diffusion, when applied on image extrapolation, often generates visible seams and repetitive patterns. MultiDiffusion creates seamless panoramas but fails to achieve global coherence across the image. In contrast, our~\SyncDiffusion{} synchronizes windows across the panorama by increasing the perceptual similarity of the denoised output predictions. This results in significantly more coherent panorama outputs.
}
\label{fig:teaser}
\end{figure}

%% file: sections/01_intro.tex
\section{Introduction}
\label{sec:intro}
\vspace{-0.5\baselineskip}
Diffusion models have recently emerged as the forefront of generative models. Recent breakthroughs in text-to-image generation such as DALL$\cdot$E 2~\cite{ramesh2022dalle2} and Stable Diffusion~\cite{rombach2022highresolution} are diffusion models trained with billions of images. Compared with GANs~\cite{goodfellow2014generative}, diffusion models offer advantages not only in producing high-quality, realistic images but also in being utilized for conditional generation setups such as inpainting~\cite{lugmayr2022repaint, nichol2021glide, saharia2022palette}, editing~\cite{meng2022sdedit, hertz2022prompt, brooks2022instructpix2pix, avrahami2023blended_latent, avrahami2022blended, nichol2021glide}, and deblurring~\cite{whang2022deblurring, chung2023dps}, with few-shot~\cite{kumari2023multiconcept, gal2023image, ruiz2023dreambooth} or even zero-shot~\cite{ramesh2021dalle, ramesh2022dalle2, lin2023magic3d, tang2023makeit3d, xu2023dream3d} adaptation. The iterative reverse diffusion process can easily incorporate diverse conditions and regularizations at each step while guiding the entire process to produce realistic images. Hence, the diffusion model, once trained on a large-scale dataset, becomes a versatile and multi-purpose pretrained model that can be applied to various tasks and applications.

Recent work has extended the versatility of pretrained diffusion models to generate arbitrary-sized images or panoramas using either sequential~\cite{avrahami2022blended} or joint~\cite{bartal2023multidiffusion} diffusion processes. Since typical image diffusion models are trained to generate fixed-sized images, creating panoramic images requires stitching multiple fixed-size images together, which can be impossible to do seamlessly without considering integration in the generation process. Two approaches have been proposed to tackle this issue. The first approach~\cite{avrahami2022blended} involves generating the final output as a sequence of inpainting starting from an initial image, where each consecutive view image is produced while fixing the overlapped region (which is referred to as \textit{image extrapolation} in their work~\cite{avrahami2022blended}).
However, this approach often struggles to seamlessly extend the given image and also tends to repeat similar patterns, resulting in unrealistic panoramas as shown in the first row of Fig.~\ref{fig:teaser}.
The other approach is joint diffusion~\cite{bartal2023multidiffusion}, which operates the reverse generative process simultaneously across multiple views while averaging the intermediate noisy images (or the noisy latent features) in the overlapped regions at each reverse process step. 
The blending of noisy latent features among the views at each denoising step can effectively generate a seamless montage of images. However, it is important to note that the content and styles of the images may vary across the views, resulting in a mixture of colorful and black-and-white images in a single panorama, as shown in the second row of Fig.~\ref{fig:teaser}. The lack of consistency occurs because the latent features of the overlapped regions are simply averaged without considering the coherence between them.

To address the limitation of previous work that produces unrealistic or incoherent montages, we present a novel synchronization module for joint diffusion, dubbed \SyncDiffusion{}. This module guides the reverse diffusion processes to achieve global coherence across different areas of the panorama image. Similar to previous guided diffusion methods~\cite{ramesh2022dalle2, lugmayr2022repaint, nichol2021glide, saharia2022palette}, our \SyncDiffusion{} guides the reverse diffusion process while adjusting the intermediate noisy images at each step. Our guidance is specifically provided as a gradient descent from a perceptual similarity loss calculated across multiple windows. Various off-the-shelf perceptual similarity losses such as LPIPS~\cite{zhang2018lpips} or Style Loss~\cite{gatys2016styleloss} 
can be utilized in our framework. However, perceptual similarity losses computed with noisy images cannot effectively guide the denoising process. Thus, we draw inspiration from the non-Markovian formulation of DDIM~\cite{song2022denoising} leveraging the prediction of the \textit{denoised output} from the current noisy image at each denoising step. We compute the perceptual similarity loss using the \textit{foreseen} denoised images at each step and then backpropagate the gradient through the noisy images.
By leveraging the synergy with a prior seamless stitching technique~\cite{bartal2023multidiffusion} based on averaging latent features at each denoising step, our joint diffusion framework demonstrates the capability to generate montages that exhibit both local seamlessness and global coherence, as shown in the last row of Fig.~\ref{fig:teaser}. This is achieved in a zero-shot manner, without the need for retraining or fine-tuning of existing diffusion models.

In our experiments on text-guided panorama generation using Stable Diffusion 2.0~\cite{rombach2022highresolution} model, the results demonstrate that our method achieves significantly higher coherence compared to previous methods. Quantitatively, as we increase the weight of the gradient descent, we observe improved coherence, measured by LPIPS~\cite{zhang2018lpips} and Style Loss~\cite{gatys2016styleloss}, while maintaining fidelity (measured by GIQA~\cite{gu2020giqa}) and compatibility with the input prompt (assessed by the CLIP score~\cite{hessel2021clipscore}). Diversity, measured by FID and KID, exhibits a trade-off with coherence, but our method still achieves much better scores compared to the baseline method. 
Our user studies confirm a significant preference for our method (66.35\%) over the previous method (33.65\%) in terms of coherence, while also suggesting superior image quality and higher prompt compatibility.
Moreover, we further demonstrate the versabtility of \SyncDiffusion{} across three plug-and-play applications: layout-guided image generation, conditional image generation and 360-degree panorama generation.

\clearpage

%% file: sections/02_related.tex
\section{Related Work}
\label{sec:related}

\vspace{-0.5\baselineskip}
\subsection{Diffusion Models}
\vspace{-0.5\baselineskip}
Diffusion probabilistic models \cite{sohldickstein2015deep, dhariwal2021diffusion, ho2020denoising, nichol2021improved} are a group of generative models that generate data by sampling from an initial Gaussian distribution and iteratively applying a denoising process (referred to as the reverse process). These methods have achieved remarkable strides in image generation \cite{rombach2022highresolution, dhariwal2021diffusion, saharia2022photorealistic, ramesh2022dalle2}, surpassing previous image generation models including GANs \cite{karras2020analyzing, brock2019large}. 
DDPM~\cite{ho2020denoising} was among the pioneering models that showcased the impressive capability of image generation through Markovian forward and reverse processes, albeit with longer computation times in generation. This limitation was addressed by DDIM~\cite{song2022denoising}, which significantly reduced the sampling time in the reverse process using a non-Markovian transition formulation conditioned on the predicted denoised data. Furthermore, LDM \cite{rombach2022highresolution} proposed incorporating the diffusion process into the latent space,
achieving state-of-the-art realism in generated images and gaining attention in the text-to-image generation.
Diffusion models have also demonstrated their applicability to diverse data modalities such as audio \cite{yang2022diffsound, huang2023makeanaudio, liu2023audioldm, ghosal2023texttoaudio}, videos~\cite{ho2022video, blattmann2023videoldm}, and 3D objects~\cite{poole2022dreamfusion, lin2023magic3d, tang2023makeit3d, xu2023dream3d}.

\vspace{-0.75\baselineskip}
\subsection{Few-Shot or Zero-Shot Adaptation of Diffusion Models}
\vspace{-0.5\baselineskip}
Building upon the remarkable generation capabilities of pretrained public text-to-image diffusion models such as Stable Diffusion~\cite{rombach2022highresolution}, recent research has introduced various methodologies for leveraging the pretrained models in diverse tasks including conditional generation, image editing, and manipulation, without the need to retrain the models from scratch. ControlNet~\cite{zhang2023adding} is an example of a method that enables the incorporation of additional conditions into existing text-to-image diffusion models through few-shot finetuning, wherein the image encoder is duplicated to handle the additional conditional image, and only a carefully selected subset of parameters is modified during the finetuning process. Custom Diffusion~\cite{kumari2023multiconcept} also introduces a similar idea of enabling few-shot tuning while keeping the majority of parameters in the neural network frozen, but with applications of finetuning the model for a particular class or concept of images.
Other previous work has also demonstrated that diffusion models can even be applied to novel tasks in a zero-shot manner. SDEdit~\cite{meng2022sdedit} was the first to show zero-shot conditional image generation using a pretrained diffusion model by dispersing noise over the conditional image and denoising it back to a real image. RePaint~\cite{lugmayr2022repaint} introduced an image inpainting idea by combining a generated foreground image and a noised background image at each time step. Similar {\em guided} diffusion ideas have also been explored for various tasks, such as image super-resolution~\cite{li2021srdiff, gao2023implicit, chung2023dps, saharia2021image}, colorization~\cite{saharia2022palette, chung2022mcg}, deblurring~\cite{whang2022deblurring, chung2023dps}, and style transfer~\cite{kwon2022diffusion, kumari2023multiconcept, gal2023image, ruiz2023dreambooth}. We propose a novel guided diffusion framework for image montage generation via joint diffusion.

\vspace{-0.75\baselineskip}
\subsection{Montage Generation via Diffusion Models}
\vspace{-0.5\baselineskip}
Panorama generation is one of the zero-shot applications of diffusion models.
Since diffusion models are trained to generate images of a specific size and on a 2D plane, stitching is required to generate panoramas or textures. Most previous methods~\cite{richardson2023texture, chen2023text2tex, avrahami2022blended, avrahami2023blended_latent} have employed inpainting-based approaches for seamless stitching. These methods extrapolate the accumulated image and fill only the missing regions to generate the panorama or texture. 
In contrast, MultiDiffusion~\cite{bartal2023multidiffusion} and DiffCollage~\cite{zhang2023diffcollage} conduct diffusion in multiple views jointly while combining noisy latent features or scores at each reverse diffusion step.
While both approaches have successfully produced continuous images, they have limitations in enforcing global coherence across the panorama or texture. 
MVDiffusion~\cite{tang2023mvdiffusion}, a concurrent work, extends multi-view diffusion to produce non-square panorama images such as 360 panorama images by leveraging pixel-wise correspondence and attention modules
However, it focuses on achieving smooth stitching, without addressing global coherence.
In this work, we propose a simple yet effective synchronization module that can be integrated into any joint diffusion process to achieve global semantic coherence.

%% file: sections/03_method.tex
\vspace{-\baselineskip}
\section{Backgrounds}
\vspace{-0.5\baselineskip}

\subsection{Diffusion Models}
\label{sec:diffusion_models}
\vspace{-0.5\baselineskip}

In this section, we provide a brief overview of the Denoising Diffusion Probabilistic Models (DDPM)~\cite{ho2020denoising} and Denoising Diffusion Implicit Models (DDIM)~\cite{song2022denoising}, which are the foundations of recent pretrained image diffusion models. The aim of DDPM is to approximate the data distribution $q(\V{x}_0)$ with a tractable model distribution $p_\theta(\V{x}_0)$, which takes the form of a \emph{Markov} chain with learned Gaussian transitions $p_\theta(\V{x}_{t-1} | \V{x}_t)$ from $p(\V{x}_T) = \C{N}(\V{x}_T; \mathbf{0}, \V{I})$:
\begin{align}
p_\theta (\V{x}_0) = \int p_\theta(\V{x}_{0:T}) d\V{x}_{1:T}, \quad \text{where} \quad p_\theta(\V{x}_{0:T}) = p(\V{x}_T) \prod_{t=1}^{T} p_\theta(\V{x}_{t-1} | \V{x}_t).
\end{align}

The parameters of the joint distribution (known as the {\em reverse process}) $\theta$ are learned by minimizing the negative evidence lower bound (ELBO):
\begin{align}
\min_\theta \mathbb{E}_{q(\V{x}_0)} \left[-\log p_\theta(\V{x}_0)\right] \leq \min_\theta \mathbb{E}_{q(\V{x}_0, \V{x}_1, \cdots, \V{x}_T)} \left[-\log p_\theta(\V{x}_{0:T}) + \log q(\V{x}_{1:T} | \V{x}_0) \right],
\label{eq:elbo}
\end{align}
where $q_\theta(\V{x}_{1:T} | \V{x}_0)$ is the {\em forward process} adding a sequence of Gaussian noise to the data while increasing the noise scale. Among the variations of the forward processes, DDPM uses the {\em variance-preserving} diffusion that parameterizes the Gaussian transitions as follows with a decreasing sequence $\alpha_{1:T} \in (0, 1]^T$:
\begin{align}
q(\V{x}_{1:T} | \V{x}_0) := \prod_{t=1}^T q(\V{x}_t | \V{x}_{t-1}), \,\, \text{where} \,\, q(\V{x}_{t} | \V{x}_{t-1}) 
:= \C{N} \left(\sqrt{\frac{\alpha_t}{\alpha_{t-1}}} \V{x}_{t-1}, \left( 1 - \frac{\alpha_t}{\alpha_{t-1}} \right)\V{I} \right).
\end{align}

The definition of the Gaussian transitions in the forward process derives the following property:
\begin{align}
q(\V{x}_{t} | \V{x}_0) := \C{N} (\V{x}_t; \sqrt{\alpha_t} \V{x}_0, (1-\alpha_t) \V{I}),
\end{align}
and thus matches the choice of the starting distribution in the reverse process (a unit Gaussian) since $q(\V{x}_{T} | \V{x}_0)$ converges to a unit Gaussian when $\alpha_T$ is set close to 0. It also allows expressing $\V{x}_t$ with $\V{x}_0$ and a unit Gaussian noise variable $\epsilon$:
\begin{align}
\V{x}_{t} = \sqrt{\alpha_t} \V{x}_0 + \sqrt{1 - \alpha_t} \epsilon, \quad \text{where} \quad \epsilon \sim \C{N}(\mathbf{0}, \V{I}).
\end{align}

In DDPM~\cite{ho2020denoising}, the Gaussian transition $p_\theta(\V{x}_{t-1} | \V{x}_t)$ for each $\V{x}_t$ in the reverse process is modeled as follows~\footnote{Note that $\beta_t$ in the DDPM~\cite{ho2020denoising} is equivalent to $1-\frac{\alpha_t}{\alpha_{t-1}}$ in our paper and DDIM~\cite{song2022denoising}.}:
\begin{align}
p_\theta (\V{x}_{t-1} | \V{x}_t) := \C{N} \left(\sqrt{\frac{\alpha_{t-1}}{\alpha_t}} \left( \V{x}_t - \frac{1}{\sqrt{1 - \alpha_t}}\left( 1 - \frac{\alpha_t}{\alpha_{t-1}} \right)\epsilon_\theta(\V{x}_t, t) \right), \sigma^2_t \V{I} \right),
\label{eq:ddpm_p}
\end{align}

where $\sigma^2_t = \frac{1-\alpha_{t-1}}{1-\alpha_t} \cdot \left(1 - \frac{\alpha_t}{\alpha_{t-1}}\right)$, and $\epsilon_\theta(\V{x}_t, t)$ is a learned function that optimizes the objective in Eq.~\ref{eq:elbo} when it maps each $\V{x}_t$ at time $t$ to a unit Gaussian noise, thus resulting in the following simplified loss:
\begin{align}
L(\epsilon_\theta) := \sum_{t=1}^T \mathbb{E}_{\V{x}_0 \sim q(\V{x}_0), \epsilon_t \sim \C{N}(\mathbf{0}, \V{I})} \left[ \| \epsilon_\theta \left( \sqrt{\alpha_t} \V{x}_0 + \sqrt{1 - \alpha_t} \epsilon_t, t \right) - \epsilon_t \|^2_2 \right].
\label{eq:loss}
\end{align}

DDIM~\cite{song2022denoising} provides a different perspective of seeing the same forward process as a {\em non-Markovian} process while taking the input data $\V{x}_0$ into consideration in {\em reversed} transitions:
\begin{align}
\begin{gathered}
q(\V{x}_{1:T} | \V{x}_0) := q(\V{x}_T | \V{x}_0) \prod_{t=2}^T q(\V{x}_{t-1} | \V{x}_{t}, \V{x}_0), \quad \text{where} \quad q(\V{x}_T | \V{x}_0) = \C{N}(\sqrt{\alpha_T}\V{x}_0, (1-\alpha_T)\V{I}) \\
\text{and} \quad \forall s < t, \quad q(\V{x}_s | \V{x}_{t}, \V{x}_0) = \C{N} \left(\sqrt{\alpha_s}\V{x}_0 + \sqrt{1 - \alpha_s - \sigma^2_t} \cdot \frac{\V{x}_t - \sqrt{\alpha_t}\V{x}_0}{\sqrt{1 - \alpha_t}}, \sigma^2_t \V{I} \right).
\end{gathered}
\end{align}

Then, each transition in the reverse process is also redefined as first predicting the {\em denoised observation} $\V{x}_0$ given each $\V{x}_t$ and then sampling $\V{x}_{t-1}$ via the conditional distribution $q(\V{x}_{t-1} | \V{x}_{t}, \V{x}_0)$:
\begin{align}
p_\theta (\V{x}_{t-1} | \V{x}_t) :=
\begin{cases}
    q(\V{x}_{t-1} | \V{x}_t, \phi_\theta(\V{x}_t, t))   & \text{if } t \geq 2\\
    \C{N}(\phi_\theta(\V{x}_t, t), \sigma^2_t\V{I})     & \text{if } t = 1,
\end{cases}
\label{eq:ddim_p}
\end{align}
where
\begin{align}
\phi_\theta(\V{x}_t, t) = \frac{1}{\sqrt{\alpha_t}} (\V{x}_t - \sqrt{1 - \alpha_t} \epsilon_\theta(\V{x}_t, t))
\label{eq:pred_denoising}
\end{align}
is the predicted denoised observation. 
The key observations of DDIM are twofold. First, the same simplified objective (Eq.~\ref{eq:loss}) can be used to find the best models $\epsilon_\theta(\V{x}_t, t)$ in Eq.~\ref{eq:pred_denoising} that minimize the negative ELBO (Eq.~\ref{eq:elbo}). This means that the DDIM reverse process can be used with a pretrained DDPM without retraining. Second, a subset of the time sequence $[1, \cdots, T]$ can be used in the reverse process of DDIM since $\V{x}_s$ for any $s < t$ can be sampled from $\V{x}_t$ via the $\V{x}_0$ prediction, enabling a significant boost in the reverse process computation.

In the rest of the paper, the operation sampling the next denoised data in the reverse process with the learned distribution $p_\theta (\V{x}_{t-1} | \V{x}_t)$ defined in either Eq.~\ref{eq:ddpm_p} (DDPM) or Eq.~\ref{eq:ddim_p} (DDIM) is denoted as:
\begin{align}
\C{S}(\V{x}_t, t, \epsilon),
\label{eq:sampler}
\end{align}
which takes a noisy data $\V{x}_t$ at timestep $t$ and a unit Gaussian noise $\epsilon \sim \C{N}(\mathbf{0}, \V{I})$ as input.

\vspace{-0.5\baselineskip}
\subsection{Joint Diffusion}
\label{sec:joint_diffusion}
\vspace{-0.5\baselineskip}

In an image diffusion model, each sample from the data distribution is either a 2D grid of per-pixel colors, or a 2D grid of latent features (as in Latent Diffusion~\cite{rombach2022highresolution}) that can be encoded from or decoded to a real image through a pretrained encoder $\C{E}$ and decoder $\C{D}$.
In the rest of the paper, the term {\em image} will thus be used to refer to either a {\em color} image or a {\em latent feature} image, unless explicitly stated otherwise.
Image diffusion models pretrained on fixed-size images cannot be used directly to produce arbitrary-size images.
MultiDiffusion~\cite{bartal2023multidiffusion} has addressed this limitation by using a multi-window joint diffusion approach. The framework integrates images generated from multiple windows seamlessly by {\em averaging} colors or features across the windows at {\em every} reverse diffusion step. For instance, consider the case of generating a panorama image $\V{z} \in \mathbb{R}^{H_z \times W_z \times D}$. The image at each window $\V{x}^{(i)} \in \mathbb{R}^{H_x \times W_x \times D}$ is a subarea of the panorama image whose union across all the windows covers the entire panorama image.
Let $\V{m}^{(i)} \in [0, 1]^{H_x \times H_x}$ denote a binary mask for the subregion in the panorama image corresponding to the $i$-th window. The function $\C{T}_{\V{z} \rightarrow i}: \mathbb{R}^{H_z \times W_z \times D} \rightarrow \mathbb{R}^{H_x \times W_x \times D}$ maps (crops) the panorama image $z$ to the $i$-th window image, while $\C{T}_{i \rightarrow \V{z}}: \mathbb{R}^{H_x \times W_x \times D} \rightarrow \mathbb{R}^{H_z \times W_z \times D}$ is its inverse function that fills the region outside of the mask $\V{m}_i$ with zeros. During the joint diffusion process running the reverse process simultaneously for each window, the noisy images from the windows ${\V{x}^{(i)}_t}$ are first averaged in the panorama space:
\begin{align}
\V{z}_{t} = \frac{\sum_i \C{T}_{i \rightarrow \V{z}} (\V{x}^{(i)}_t)}{\sum_i \V{m}^{(i)}},
\label{eq:multi_diffusion}
\end{align}
and then, the resulting combined noisy image $\V{z}_{t}$ is cropped again for each window $\V{x}^{(i)}_t = \C{T}_{\V{z} \rightarrow i}(\V{z}_t)$, modifying the noisy image at each window with a inter-window regularization.

\begin{figure}[h!]
\centering
{
    \setlength{\fboxrule}{1pt}
    \setlength{\fboxsep}{0pt}
    \fcolorbox{red}{black}{\includegraphics[width=0.48\textwidth]{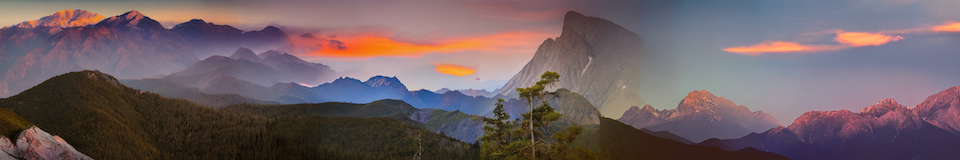}}
    \fcolorbox{black}{black}{\includegraphics[width=0.48\textwidth]{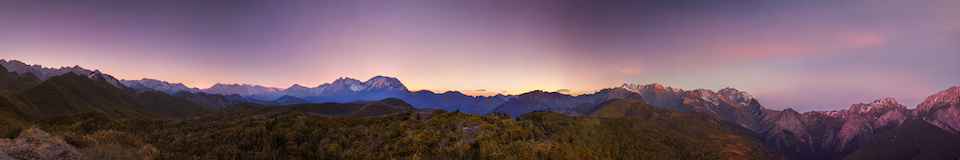}}
}
\vspace{-0.5\baselineskip}
\caption{
Panoramas generated by MultiDiffusion~\cite{bartal2023multidiffusion} (left) and our~\SyncDiffusion{} (right), with a prompt \textit{``A photo of a mountain range at twilight''}.
MultiDiffusion often combines various scenes, such as mountains with trees and snow, and even awkwardly blends them. In contrast, \SyncDiffusion{} generates panoramas that are significantly more coherent.
}
\label{fig:md_failure}
\vspace{-0.5\baselineskip}
\end{figure}

\vspace{-0.5\baselineskip}
\section{SyncDiffusion}
\label{sec:method}
\vspace{-0.5\baselineskip}

\begin{figure}[t]
  \includegraphics[width=\textwidth]{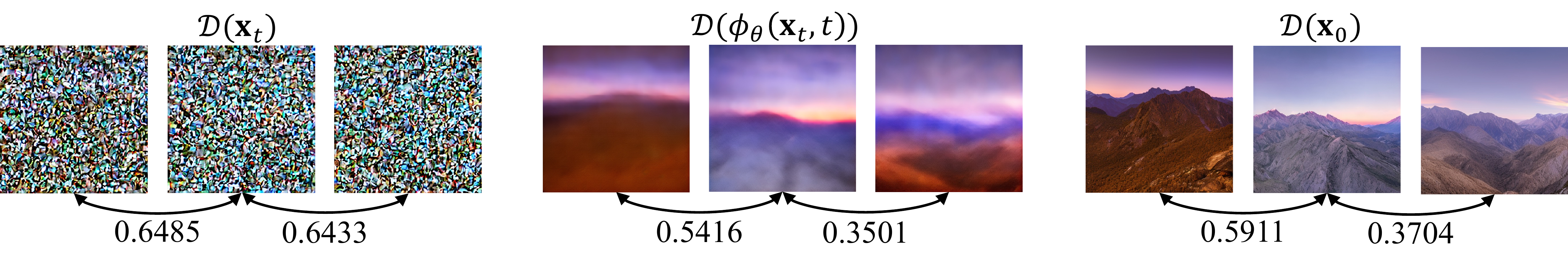}
  \vspace{-1.5\baselineskip}
  \caption{
  LPIPS~\cite{zhang2018lpips} scores computed across the noisy images $\C{D}(\V{x}_t)$ at the intermediate step ($t=45$ out of $50$) of the reverse process (left), the {\em predicted} denoised images $\C{D}(\phi(\V{x}_t, t))$ at the same timestep $t$ (middle), and the final generated images $\C{D}(\V{x}_0)$ at timestep $t=0$ (right). The indistinguishable noisy images yield similar LPIPS scores among them, whereas the predicted denoised images, which closely resemble the final outputs even at the beginning of the denoising process, exhibit LPIPS scores that align with those of the final generated images.
  This indicates that the predicted denoised images can provide meaningful guidance for producing coherent panoramas in the diffusion process.}
  \label{fig:lpips_example}
\end{figure}

\begin{algorithm}[!t]
\caption{Pseudocode of one-time denoising in~\SyncDiffusion{}.}
\label{alg:method} 
{
    \footnotesize
    \KwParam{$w$\tcp*{Gradient descent weight.}}
    \KwInput{$\{ \V{x}^{(i)}_{t} \}_{i=0 \cdots N-1}$\tcp*{Noisy images at timestep $t$.}}
    \KwOutput{$\{ \V{x}^{(i)}_{t-1} \}_{i=0 \cdots N-1}$\tcp*{Noisy images at timestep $t-1$.}}
    
    \SetKwFunction{FSD}{SyncDiffusion}
    \SetKwFunction{FMD}{MultiDiffusion}
    \SetKwFunction{FMain}{DenoisingOneStep}
    
    \SetKwProg{Fn}{Function}{:}{}
    \Fn{\FSD{$\{ \V{x}^{(i)}_{t} \}$}}{
        $\V{\hat{x}}^{(i)}_t \leftarrow \V{x}^{(i)}_t$\tcp*{The anchor window at index 0 is not changed.}
        \For{$i = 1, \dots, N-1$} {
            $\V{\hat{x}}^{(i)}_t \leftarrow \V{x}^{(i)}_t - w\nabla_{\V{x}^{(i)}_t} \C{L}\left(\C{D}(\phi_\theta(\V{x}^{(i)}_t, t)), \C{D}(\phi_\theta(\V{x}^{(0)}_t, t))\right)$\tcp*{Gradient descent (Eq.~\ref{eq:gradient_descent})}
        }
        \KwRet $\{ \V{\hat{x}}^{(i)}_{t} \}$\;
    }
    \SetKwProg{Fn}{Function}{:}{}
    \Fn{\FMD{$\{ \V{\tilde{x}}^{(i)}_{t} \}$}}{
        $\V{z}_{t} \leftarrow \frac{\sum_i \C{T}_{i \rightarrow \V{z}} (\V{\tilde{x}}^{(i)}_{t})}{\sum_i \V{m}^{(i)}}$\tcp*{Averaging in the global space (Eq.~\ref{eq:multi_diffusion}).}
        \For{$i = 0, \dots, N-1$} {
            $\V{x}^{(i)}_{t} \leftarrow \C{T}_{\V{z} \rightarrow i}(\V{z}_{t})$\;
        }
        \KwRet $\{ \V{x}^{(i)}_{t} \}$\;
    }
    \SetKwProg{Fn}{Function}{:}{}
    \Fn{\FMain{$\{ \V{x}^{(i)}_{t} \}$}}{
        $\{ \V{\hat{x}}^{(i)}_{t} \} \leftarrow $~\FSD{$\{ \V{x}^{(i)}_{t} \}$}\;
        \For{$i = 0, \dots, N-1$} {
            $\V{\tilde{x}}^{(i)}_{t-1} \leftarrow \C{S}(\V{\hat{x}}^{(i)}_t, t, \epsilon)$\tcp*{Sampling the next denoised data (Eq.~\ref{eq:sampler}).}
        }
        $\{ \V{x}^{(i)}_{t-1} \} \leftarrow$~\FMD{$\{ \V{\tilde{x}}^{(i)}_{t-1} \}$}\;
        \KwRet $\{ \V{x}^{(i)}_{t-1} \}$\;
    }
}
\end{algorithm}

While MultiDiffusion~\cite{bartal2023multidiffusion} can generate seamless panorama images from joint diffusion, it often fails to produce coherent and realistic montages.
The left image in Fig.~\ref{fig:md_failure} demonstrates that the resulting image often oddly combines various scenes, such as mountains with trees and snow. Also, the blending occasionally fails to merge them in a realistic manner, as shown in the figure where distant objects are connected to closer objects.
This incoherence issue in MultiDiffusion arises due to two main reasons. Firstly, the averaging operation only aligns the colors or features in the overlapped regions but does not match the {\em content} or {\em style} of the images.
Secondly, it only enforces {\em adjacent} views to influence each other, and thus global coherence between distant windows cannot be achieved.

To address this problem, we introduce a module called~\SyncDiffusion{} which enables the generation of coherent montages, as shown on the right in Fig.~\ref{fig:md_failure}. This module can be easily integrated into an existing joint diffusion framework.
Similar to MultiDiffusion, our \SyncDiffusion{} module updates the noisy image at every step of the reverse diffusion process. In contrast to averaging the colors or latent features in the overlapped regions, however, \SyncDiffusion{} employs the backpropagation of gradients from a perceptual similarity loss computed across the windows to perform the update.
The perceptual similarity loss, denoted as $\C{L}$, can utilize any off-the-shelf loss function for perceptual similarity, such as LPIPS~\cite{zhang2018lpips} and Style Loss~\cite{gatys2016styleloss}.
To facilitate efficient computation, we designate an {\em anchor} window with an index of $0$. For each view's noisy color image $\C{D}(\V{x}^{(i)}_t)$ and the anchor window's noisy color image $\C{D}(\V{x}^{(0)}_t)$ (where the decoder $\C{D}$ can be treated as an identity function if the given diffusion model operates in image space rather than latent space), one can measure the coherence using the images and conduct gradient descent through $\V{x}^{(i)}_t$:
\begin{align}
\V{\hat{x}}^{(i)}_t = \V{x}^{(i)}_t - w\nabla_{\V{x}^{(i)}_t} \C{L}\left(\C{D}(\V{x}^{(i)}_t), \C{D}(\V{x}^{(0)}_t)\right),
\label{eq:gradient_descent_x_t}
\end{align}
where $w$ is the weight of the gradient descent. However, the coherence measured with the {\em noisy} images cannot provide meaningful guidance.
Fig.~\ref{fig:lpips_example} shows examples where the left three images are the intermediate noisy images $\C{D}(\V{x}_t)$ at timestep $t = 45$ out of a total of $50$ timesteps in the DDIM reverse process. Note that the LPIPS scores among the noisy images are indistinguishable.
Hence, similar to the DDIM reverse process, we utilize the {\em foreseen denoised} observation of each noisy data $\phi_\theta(\V{x}^{(i)}_t, t)$ in Eq.~\ref{eq:pred_denoising}. We measure the coherence not with the current noisy color images $\{ \C{D}(\V{x}^{(i)}_t) \}$ but with the predicted denoised color images $\{ \C{D}(\phi_\theta(\V{x}^{(i)}_t, t)) \}$ with the timestep $t$ and perform the backpropagation of the gradient through $\V{x}^{(i)}_t$, resulting in the following updated formulation:
\begin{align}
\V{\hat{x}}^{(i)}_t = \V{x}^{(i)}_t - w\nabla_{\V{x}^{(i)}_t} \C{L}\left(\C{D}(\phi_\theta(\V{x}^{(i)}_t, t)), \C{D}(\phi_\theta(\V{x}^{(0)}_t, t))\right).
\label{eq:gradient_descent}
\end{align}
In Fig.~\ref{fig:lpips_example}, the middle three images depict the predicted denoised images $\C{D}(\phi_\theta(\V{x}_t, t))$ at timestep $t = 45$, which closely resemble the final generated images $\C{D}(\V{x}_0)$ at timestep $t = 0$, even during the initial stages of the reverse diffusion process. Therefore, the LPIPS scores among the predicted denoised images also match those of the generated images, providing meaningful guidance for maintaining coherence.
During each denoising step in the joint reverse process, we apply this update to the noisy images for all windows $\{ \V{x}^{(i)}_t \}$, and sample the one-step denoised images. MultiDiffusion is also applied to average the sampled images at the end.
Refer to Alg.~\ref{alg:method} for detailed pseudocode.

%% file: sections/04_results.tex
\vspace{-0.5\baselineskip}
\section{Results}
\label{sec:results}

\vspace{-0.5\baselineskip}
\subsection{Text-Guided Panorama Generation}
\vspace{-0.5\baselineskip}

\begin{figure*}[t!]
    \centering
    \footnotesize{
        \renewcommand{\arraystretch}{0.0}
        \setlength{\tabcolsep}{0.0em}
        \setlength{\fboxrule}{0.0pt}
        \setlength{\fboxsep}{0pt}
        
        \begin{tabularx}{\textwidth}{>{\centering\arraybackslash}m{0.07\textwidth} >{\centering\arraybackslash}m{0.465\textwidth} >{\centering\arraybackslash}m{0.465\textwidth}}
        & \textit{``A photo of a city skyline at night''} & \textit{``A photo of a mountain range at twilight''}\\
        \scriptsize{BLD~\cite{avrahami2023blended_latent}} & 
        \includegraphics[width=0.46\textwidth]{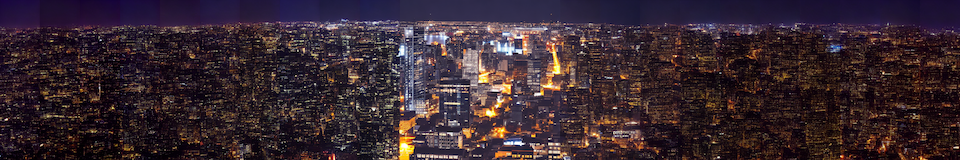} &
        \includegraphics[width=0.46\textwidth]{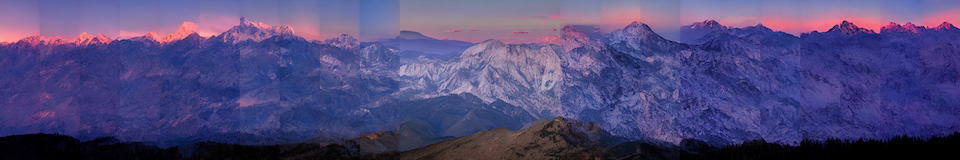} \\
        \scriptsize{MD~\cite{bartal2023multidiffusion}} & 
        \includegraphics[width=0.46\textwidth]{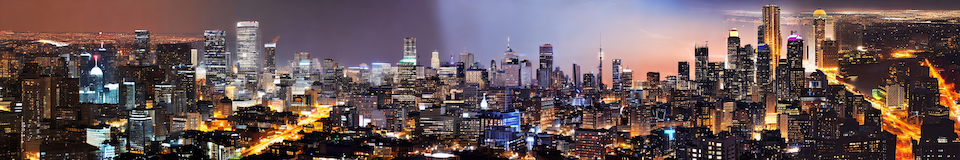} &
        \includegraphics[width=0.46\textwidth]{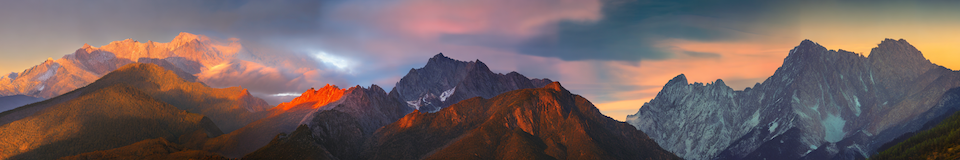} \\
        \scriptsize{\makecell{Sync\\Diffusion}} & 
        \framebox{\includegraphics[width=0.46\textwidth]{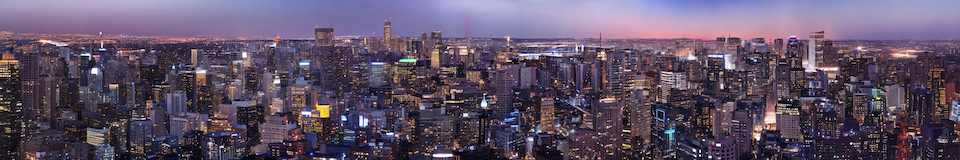}} &
        \framebox{\includegraphics[width=0.46\textwidth]{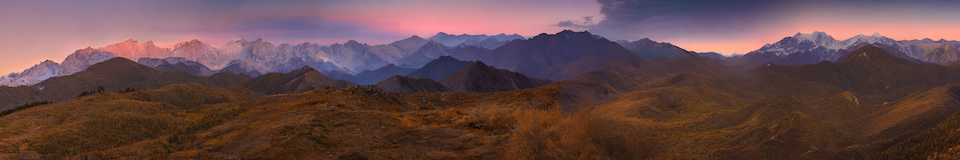}} \\
        \midrule
        & \scalebox{.85}[1.0]{\textit{``A cartoon panorama of spring summer beautiful nature''}} & \textit{``A natural landscape in anime style illustration''}\\
        \scriptsize{BLD~\cite{avrahami2023blended_latent}} & 
        \includegraphics[width=0.46\textwidth]{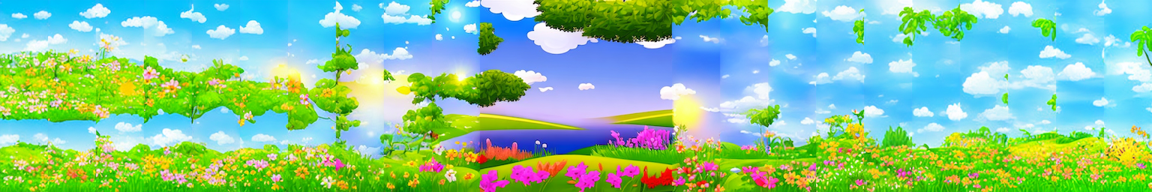} &
        \includegraphics[width=0.46\textwidth]{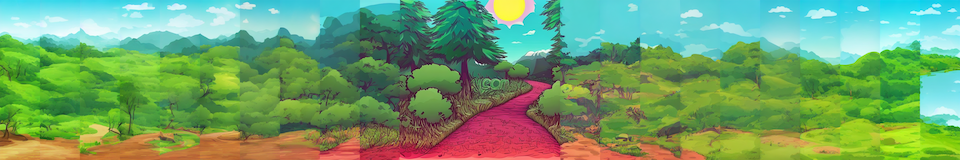} \\
        \scriptsize{MD~\cite{bartal2023multidiffusion}} & 
        \includegraphics[width=0.46\textwidth]{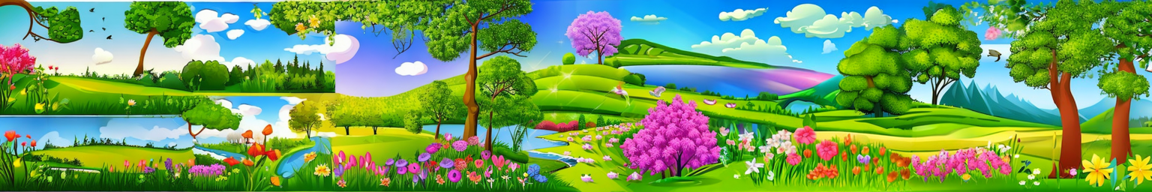} &
        \includegraphics[width=0.46\textwidth]{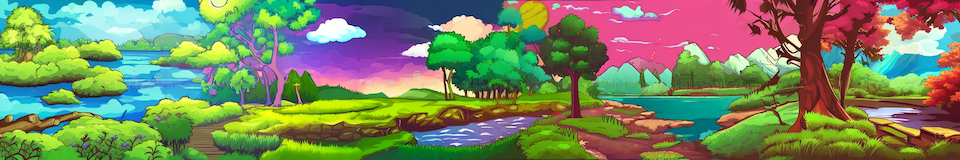} \\
        \scriptsize{\makecell{Sync\\Diffusion}} & 
        \framebox{\includegraphics[width=0.46\textwidth]{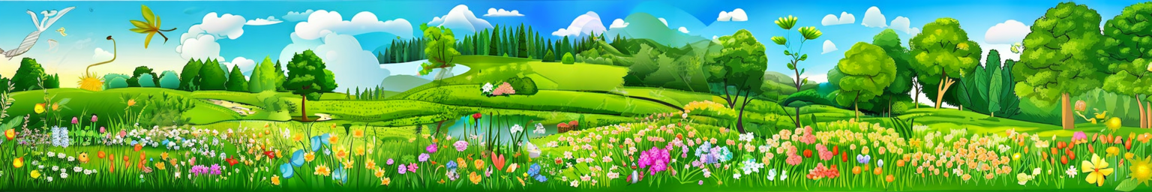}} &
        \framebox{\includegraphics[width=0.46\textwidth]{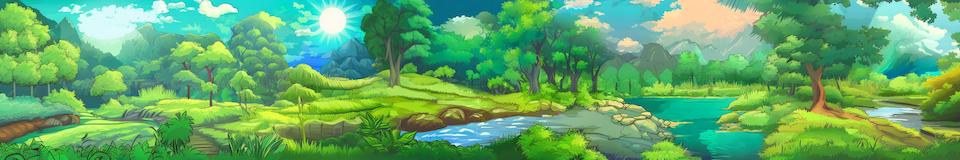}} \\
        \end{tabularx}
    }
    \vspace{-0.5\baselineskip}
    \caption{Qualitative comparisons. Blended Latent Diffusion~\cite{avrahami2023blended_latent} (BLD, the first row of each case) tends to exhibit visible seams and repetitive patterns. MultiDiffusion~\cite{bartal2023multidiffusion} (MD, the second row of each case) generates seamless results but lacks coherence, such as blending a sunset sky with a blue sky (top right), and displaying a combination of purple, pink, and blue backgrounds (bottom). In contrast, our~\SyncDiffusion{} (the last in each case) produces seamless panoramas with significantly improved coherence. Best view in zoom and color.}
    \label{fig:qualitative}
\end{figure*}

In our experiments, we generate panorama images using our~\SyncDiffusion{} method and the pretrained Stable Diffusion 2.0 \cite{rombach2022highresolution} model.
Stable Diffusion model operates in a latent space of $\mathbb{R}^{64\times64\times4}$ and generates images of $\mathbb{R}^{512\times512\times3}$.
We generate panorama images of resolution $512 \times 3072$ ($64\times384$ in the latent space), where the width is six times the width of the output of Stable Diffusion. Each window $\V{x}^{(i)}$ has an image resolution of $512 \times 512$, with a stride of $128$ pixels along the width in the image space which is equivalent to stride 16 in the latent space, resulting in a total 21 windows to operate diffusion processes jointly.
We use six text prompts from MultiDiffusion~\cite{bartal2023multidiffusion} (see Sec.~\ref{sup:more_qualitative}) and generate 500 panoramas per prompt.
For the gradient descent weight $w$ (Eq.~\ref{eq:gradient_descent}), we experiment with various initial values while applying a weight decay with a rate of $0.95$. We also set the center window as the anchor window with an index of $0$.

\vspace{-0.75\baselineskip}
\paragraph{Baselines}
We compare our~\SyncDiffusion{} with previous methods that generate panoramas using a pretrained diffusion model. Blended Latent Diffusion~\cite{avrahami2023blended_latent} is an inpainting-based method that extrapolates a single window image. MultiDiffusion~\cite{bartal2023multidiffusion}, served as the base of our framework, is a special case of our method when the weight of gradient descent $w$ is $0$. The same Stable Diffusion 2.0 model is used for all the methods for a fair comparison.

\vspace{-0.75\baselineskip}
\paragraph{Evaluation Metrics}
We utilize a range of metrics to assess the coherence, fidelity, diversity, and compatibility of the output panoramas with the input prompt.
\begin{itemize}[leftmargin=*,noitemsep,topsep=0pt]
\item (\underline{Coherence}) \textbf{Intra-LPIPS}~\cite{zhang2018lpips} and \textbf{Intra-Style-L}~\cite{gatys2016styleloss}:
To assess the coherence of the generated panoramas, we introduce two metrics. Intra-LPIPS and Intra-Style-L, which are computed as the averages of LPIPS~\cite{zhang2018lpips} and Style Loss~\cite{gatys2016styleloss}, respectively, between a pair of non-overlapping window images from the same panorama.
Specifically, we divide the panorama into 6 windows, each with dimensions of $512\times512$ and then compute the average of LPIPS and Style Loss across the 15 combinations of these cropped views. To provide a reference for the scale of these values, we generate 500 single-window-size images using the same Stable Diffusion model and compute the LPIPS and Style Loss for randomly selected 1,000 pairs of these reference images.
\item (\underline{Fidelity}) \textbf{Mean-GIQA}~\cite{gu2020giqa}:
GIQA quantifies the fidelity of {\em individual} images by calculating the inverse of the distance between a query image and a reference set in a feature space. Mean-GIQA is computed by taking a {\em single} random crop of each panorama in $512\times512$ size and computing the average GIQA score from each cropped image to the reference set of images mentioned above.
\item (\underline{Fidelity \& Diversity}) \textbf{FID}~\cite{heusel2018fid} and \textbf{KID}~\cite{kid2018}:
FID~\cite{heusel2018fid} and KID~\cite{kid2018} are used to measure both fidelity and diversity. Both of them are measured with the aforementioned randomly cropped images and the set of reference images.
\item (\underline{Compatibility with the Input Prompt}) \textbf{Mean-CLIP-S}~\cite{hessel2021clipscore}:
The compatibility with the input prompt is assessed using the mean of CLIP scores~\cite{hessel2021clipscore}, denoted as Mean-CLIP-S. This metric is calculated using the same set of cropped images and the input prompt.
\end{itemize}

\vspace{-0.75\baselineskip}
\paragraph{Qualitative Comparisons}
Fig.~\ref{fig:qualitative} showcases qualitative comparisons between our method and the baseline methods. Here, we show the results of our method generated with a weight parameter of $w=20$. Blended Latent Diffusion~\cite{avrahami2023blended_latent} often exhibits visible seams due to the sequential inpainting scheme and produces repetitive patterns in the extrapolation, as illustrated by the mountains in the second case of the top row and the flowers and trees in both cases of the bottom row.
MultiDiffusion~\cite{bartal2023multidiffusion} achieves seamless outputs, although it often produces incoherent outputs, such as mixing a sunset sky with a blue sky, as shown in the second cases of the first and second rows, and pink and purple backgrounds with blue backgrounds in the two cases of the bottom row.
Our~\SyncDiffusion{} generates visually and semantically more coherent panoramas with all the prompts. More qualitative comparisons are provided in the \textbf{Appendix (\ref{sup:more_prompts}, \ref{sup:more_qualitative})}.

\vspace{-0.75\baselineskip}
\paragraph{Quantitative Results}

\begin{figure}[h!]
    \centering
    \includegraphics[width=0.8\textwidth]{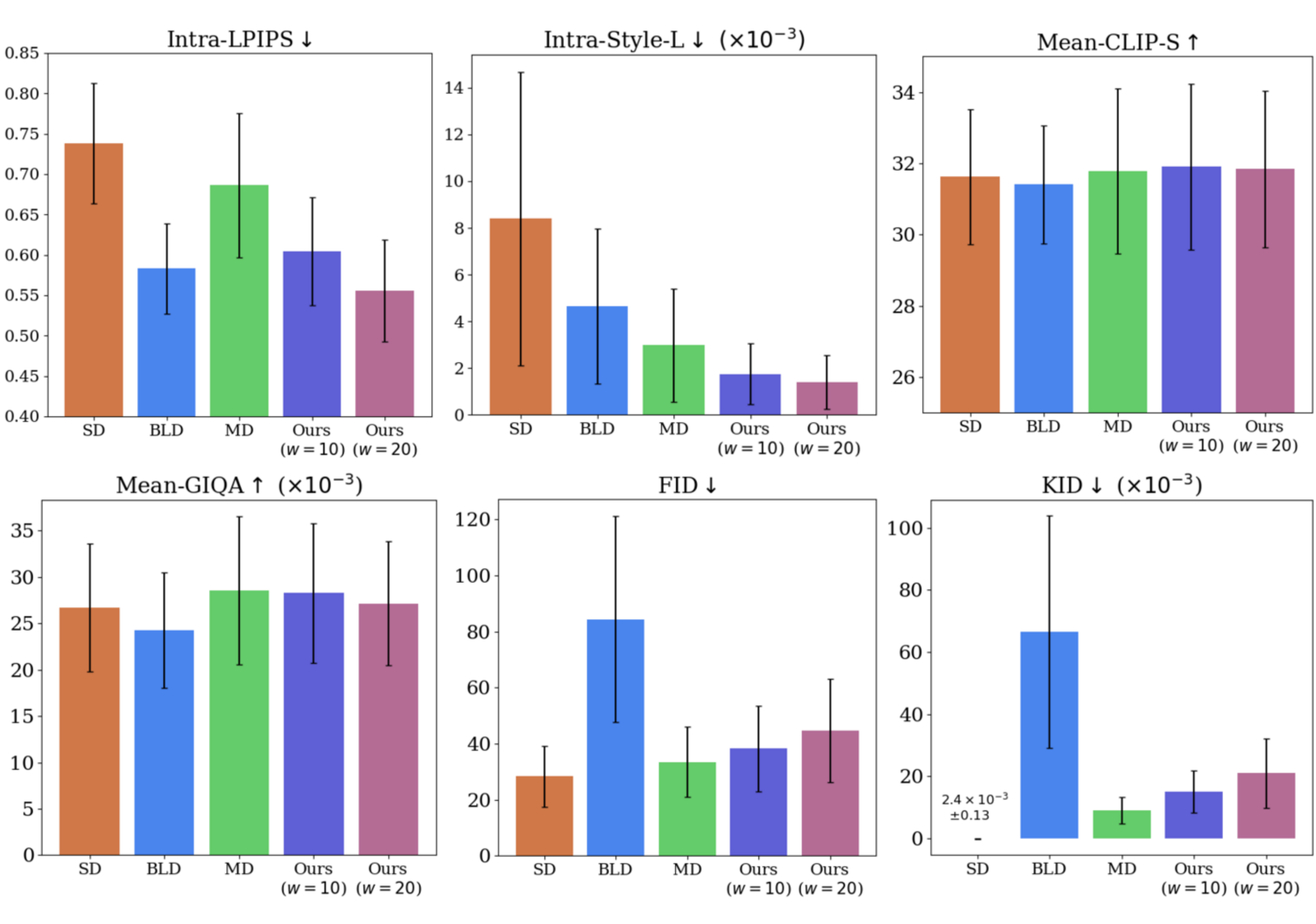}
    \label{fig:main-graph}
    \vspace{-\baselineskip}
    \caption{Quantitative results. MultiDiffusion~\cite{bartal2023multidiffusion} (MD) can be considered as a special case of our method when the gradient descent weight $w$ is set to 0. As $w$ increases, coherence (Intra-LPIPS and Intra-Style-L) improves while maintaining the compatibility with the input prompt (Mean-CLIP-S) and fidelity (Mean-GIQA). There is a trade-off between coherency and diversity, as indicated by the FID and KID results. Note that the FID and KID of our method are still significantly lower than those of Blended Latent Diffusion~\cite{avrahami2023blended_latent} (BLD). SD (Stable Diffusion) is the score with the reference set images. Refer to the text for the details.
    }
\end{figure}

Fig.~\ref{fig:main-graph} presents a quantitative comparison among the methods. Our method's results are displayed using two different gradient descent weights, $w=10$ and $20$. MultiDiffusion~\cite{bartal2023multidiffusion} is also the case when $w=0$ in our framework. For results obtained with different weights, please refer to the \textbf{Appendix (\ref{sup:more_quantitative})}. The color bars in the plots indicate the average scores across the six prompts, while the black lines depict the standard deviation. Note that as the gradient weight increases, both Intra-LPIPS and Intra-Style-L decrease. When $w=20$, the Intra-LPIPS and Intra-Style-L of our method are approximately 3/4 and 1/6 of those computed with the reference set images (referred to as SD, Stable Diffusion), respectively, indicating significantly higher coherence. Moreover, the Mean-CLIP-S and Mean-GIQA scores are comparable to those computed with the reference set, meaning that the compatibility with the input prompt and fidelity are not compromised by our diffusion synchronization.
The results of FID and KID demonstrate the trade-off between coherence and diversity. As the gradient descent weight $w$ increases, FID and KID also increase slightly, although they are still much lower compared to Blended Latent Diffusion~\cite{avrahami2023blended_latent}.
This implies that for certain images, it is more difficult to find coherent images. In the \textbf{Appendix (\ref{sup:diff_resolution})}, we substantiate this claim with the results of shorter generated panoramas.
Blended Latent Diffusion results in low Intra-LPIPS due to its tendency to repeat similar patterns, but it leads to low Mean-GIQA and very high FID and KID, indicating a significant degradation in fidelity.

\input{tables/01_main_quantitative}
\vspace{-0.75\baselineskip}
\paragraph{User Study}
We conducted three user studies to further evaluate the coherence, image quality and prompt compatibility of the generated panoramas, respectively. Following Ritchie~\cite{mturk-2afc}, participants were presented with panorama images generated by both MultiDiffusion~\cite{bartal2023multidiffusion} and our~\SyncDiffusion{} methods (with $w=20)$. 
They were then asked to choose one of them by answering the question: \texttt{Which one appears a more coherent panorama image to you?} \textbf{(Coherence)}, \texttt{Which one is of higher quality?} \textbf{(Image Quality)}, or \texttt{Which one best matches the shared caption?} \textbf{(Prompt Compatibility)}. We collected 25 responses each, including 5 vigilance tasks, from 100 participants for each user study. The results in Tab.~\ref{tab:user_study} affirm that human evaluators perceive \SyncDiffusion{} as producing more coherent results compared to MultiDiffusion, while also demonstrating superior image quality and higher prompt compatibility. Refer to the \textbf{Appendix (\ref{sup:user_study})} for detailed setups for the user study.

\begin{table}[h!]
    \centering
    {\small
    \setlength{\tabcolsep}{0.2em}
    \begin{tabularx}{0.90\linewidth}{>{\centering}m{0.15\textwidth}>{\centering\arraybackslash}m{0.23\textwidth}>{\centering\arraybackslash}m{0.23\textwidth}>{\centering\arraybackslash}m{0.23\textwidth}}
    \toprule
    \multicolumn{1}{c}{} & \multicolumn{1}{c}{Coherence ($\%$)} & \multicolumn{1}{c}{Image Quality ($\%$)} & \multicolumn{1}{c}{Prompt Compatibility ($\%$)} \\
    \midrule
    MultiDiffusion~\cite{bartal2023multidiffusion} & 33.65 & 42.81 & 40.50 \\ %
    \SyncDiffusion{}                               & \textbf{66.35} & \textbf{57.19} & \textbf{59.50} \\ %
    \bottomrule
    \end{tabularx}
    \vspace{0.5\baselineskip}
    \caption{User study results.}
    \label{tab:user_study}
    }
\end{table}

\vspace{-2.0\baselineskip}
\subsection{Additional Applications of \SyncDiffusion{}}
\vspace{-0.5\baselineskip}
We further demonstrate the versatility of \SyncDiffusion{} through three additional plug-and-play applications: layout-guided image generation, conditional image generation and 360-degree panorama generation.

\vspace{-0.75\baselineskip}
\paragraph{Layout-Guided Image Generation}
Plugging \SyncDiffusion{} into the layout-to-image pipeline in MultiDiffusion~\cite{bartal2023multidiffusion} leads to a notable enhancement in the global coherence as displayed in Fig.~\ref{fig:applications}-(A).
While MultiDiffusion (middle row) generates an unnatural image with incoherent background around the house and the bear, our method produces a natural image with a globally coherent background.

\vspace{-0.75\baselineskip}
\paragraph{Conditional Image Generation}
When integrated with ControlNet~\cite{zhang2023adding}, \SyncDiffusion{} extends the conditional image generation to arbitrary resolutions. Let $c \in \mathbb{R}^{H_z \times W_z \times 3}$ denote an input condition and $\psi_{c \rightarrow i}: \mathbb{R}^{H_z \times W_z \times 3} \rightarrow \mathbb{R}^{H_x \times W_x \times 3}$ be a mapping from $c$ to the $i$-th cropped condition corresponding to the window $\V{x}^{(i)}_t$. 
We define conditional \SyncDiffusion{} by substituting $\phi_\theta(\V{x}_t, t)$ in Alg.~\ref{alg:method} with $\phi_\theta(\V{x}_t, t, c^{(i)})$, where $c^{(i)} := \psi_{c \rightarrow i} (c)$.
Fig.~\ref{fig:applications}-(B) illustrates that the combination of ControlNet and \SyncDiffusion{} generates coherent panoramas while reflecting the given condition Canny edge map (top row).

\vspace{-0.75\baselineskip}
\paragraph{360-degree Panorama Generation}
We further plug \SyncDiffusion{} into MVDiffusion~\cite{tang2023mvdiffusion}, a concurrent work that generates 360-degree panoramas from text prompts via multi-view diffusion. As shown Fig.~\ref{fig:applications}-(C), our~\SyncDiffusion{} distinctly improves the global coherence of the generated panorama. The increase in coherence becomes more apparent when comparing perspective views from different angles. While View 1 and View 2 from the vanilla MVDiffusion (top row) seem to be from two different rooms, with our method the generated images better depict two views from the same room (bottom row).

\vspace{-0.5\baselineskip}
\paragraph{Limitations}
While our~\SyncDiffusion{} module can significantly enhance the coherence of generated panoramas, it relies on appropriate input prompts to achieve realistic results, as illustrated in Fig.~\ref{fig:limitations}. Also, the~\SyncDiffusion{} module that includes a forward pass through the neural network and gradient descent computation introduces additional computational overhead.

\vspace{-0.75\baselineskip}
\paragraph{Appendix}
Due to space constraints, we present the following additional results in the Appendix:
more qualitative comparisons with various prompts (\ref{sup:more_prompts}, \ref{sup:more_qualitative}), 
details about the quantitative evalution (\ref{sup:more_quantitative}),
evaluation on different resolutions (\ref{sup:diff_resolution}),
results with Style Loss~\cite{gatys2016styleloss} as the perceptual loss (\ref{sup:style_loss}), 
an ablation study using Eq.~\ref{eq:gradient_descent_x_t} instead of Eq.~\ref{eq:gradient_descent} (\ref{sup:x_t_ablation}),
an analysis of computation time (\ref{sup:sampling_time_analysis}),
and details on the user study (\ref{sup:user_study}).

%% file: sections/05_conclusion.tex
\begin{figure}[t!]
    \centering
    \includegraphics[width=\textwidth]{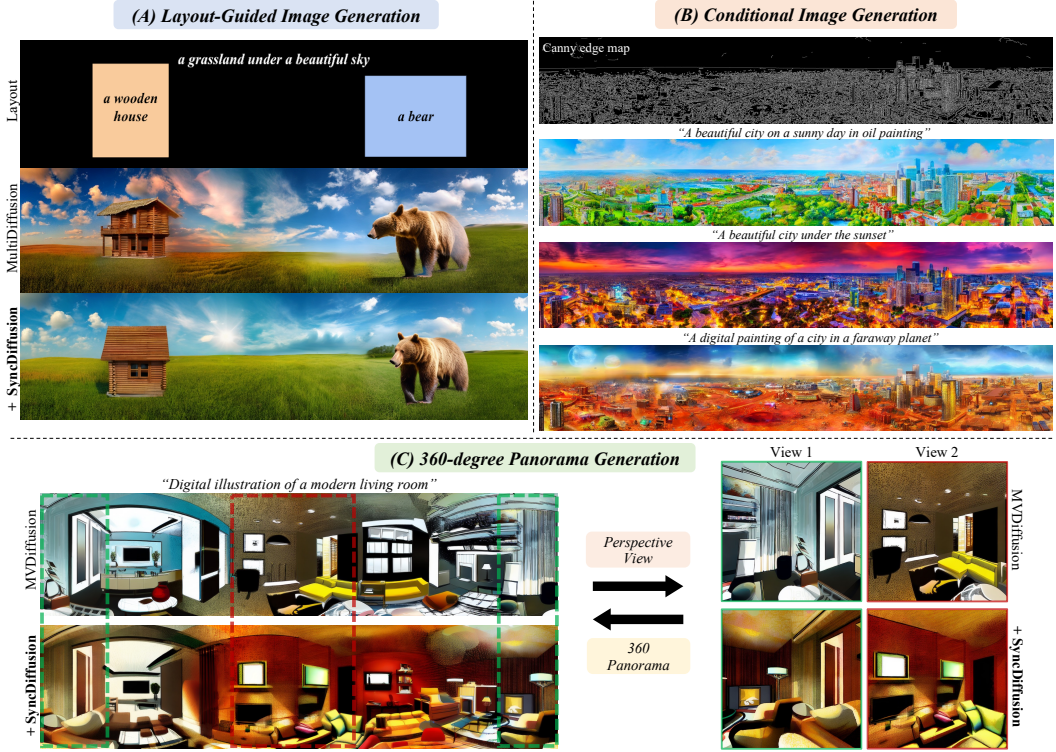}
    \vspace{-1.5\baselineskip}
    \caption{
    Plug-and-play applications of \SyncDiffusion{}.}
    \label{fig:applications}
\end{figure}

\vspace{-0.5\baselineskip}
\begin{figure*}[h!]
    \centering
    \footnotesize{
        \renewcommand{\arraystretch}{0.0}
        \setlength{\tabcolsep}{0.0em}
        \setlength{\fboxrule}{0.0pt}
        \setlength{\fboxsep}{0pt}
        \begin{tabularx}{\textwidth}{>{\centering\arraybackslash}m{0.5\textwidth} >{\centering\arraybackslash}m{0.5\textwidth}}
        \textit{``A red sports car''} & \textit{``A fancy hotel room''}\\
        \includegraphics[width=0.49\textwidth]{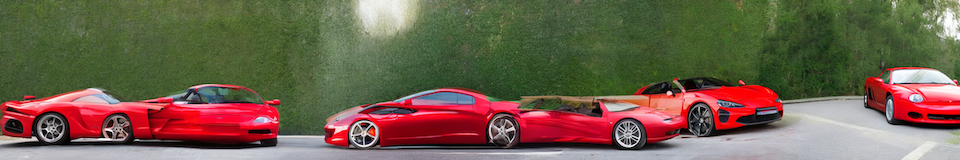} &
        \includegraphics[width=0.49\textwidth]{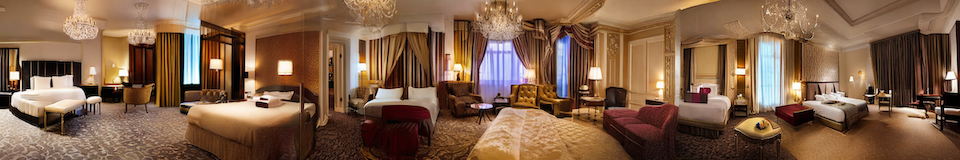}
        \end{tabularx}
    }
    \vspace{-0.5\baselineskip}
    \caption{Our failure cases. A suitable input prompt is required to generate realistic panoramas.}
    \label{fig:limitations}
\end{figure*}

\vspace{-\baselineskip}
\section{Conclusion}
\label{sec:conclude}
\vspace{-0.75\baselineskip}
We presented~\SyncDiffusion{}, a diffusion synchronization module designed to generate coherent montages through joint diffusions. Using a pretrained diffusion model, we propose guiding the reverse process by updating the noisy images at each intermediate step using gradient descent. This update is based on a perceptual similarity loss calculated with the predictions of the denoised images.
Moreover, the idea of \SyncDiffusion{} can be applied to generating textures for 3D models. We plan to investigate such possibilities in future work.

\vspace{-0.75\baselineskip}
\paragraph{Potential Negative Societal Impacts}
Image generative models can potentially generate deepfakes, images resembling copyrighted material, biased or discriminatory images, and harmful outputs. Future research is needed to advance the detection of manipulated content and establish societal barriers to protect intellectual property.

\acksection
\vspace{-0.75\baselineskip}
We thank Juil Koo for valuable discussions on diffusion models and Eunji Hong for help in conducting user studies.
This work was partially supported by the NRF grant (RS2023-00209723) and IITP grants (2019-0-00075, 2022-0-00594, RS-2023-00227592) funded by the Korean government (MSIT), the Technology Innovation Program (20016615) funded by the Korean government (MOTIE), grants from ETRI, KT, NCSOFT, and Samsung Electronics, and computing resource support from KISTI.

%% file: sections/06_supp.tex
\ifpaper
  \newcommand{\refofpaper}[1]{\unskip}
  \newcommand{\refinpaper}[1]{\unskip}
\else
  \makeatletter
  \newcommand{\manuallabel}[2]{\def\@currentlabel{#2}\label{#1}}
  \makeatother
  \manuallabel{sec:intro}{1}
  \manuallabel{sec:related}{2}
  \manuallabel{sec:diffusion_models}{3.1}
  \manuallabel{sec:joint_diffusion}{3.2}
  \manuallabel{sec:method}{4}
  \manuallabel{sec:results}{5}

  \manuallabel{alg:method}{1}
  \manuallabel{eq:gradient_descent}{13}

  \manuallabel{fig:teaser}{1}
  \manuallabel{fig:main-graph}{5}

  \manuallabel{eq:gradient_descent}{14}
  \manuallabel{eq:gradient_descent_x_t}{13}
  
  \newcommand{\refofpaper}[1]{of the main paper}
  \newcommand{\refinpaper}[1]{in the main paper}
\fi

In this supplementary document, we first show more qualitative comparisons with various prompts in Sec.~\ref{sup:more_prompts}. Sec.~\ref{sup:more_quantitative} includes a detailed quantitative evaluation of our method with different gradient descent weights ($w=0, 5, 10, 15$, and $20$). Sec.~\ref{sup:diff_resolution} shows quantitative evaluation of our method on generating panoramas of different resolutions. In Sec.~\ref{sup:style_loss}, we show the comparisons of our method with different perceptual similarity loss functions. Sec.~\ref{sup:x_t_ablation} shows an ablation study result substituting Eq.~\ref{eq:gradient_descent}~\refinpaper{} with Eq.~\ref{eq:gradient_descent_x_t}. Sec.~\ref{sup:sampling_time_analysis} analyzes the computation time of \SyncDiffusion{}. Sec.~\ref{sup:user_study} explains the details of our user study. Lastly, Sec.~\ref{sup:more_qualitative} provides additional qualitative comparisons.

\input{sections/supp_more_prompts}

\newpage

\input{sections/supp_more_quantitative}

\input{sections/supp_diff_resolution}

\vspace{-\baselineskip}
\subsection{Results of \SyncDiffusion{} with Style Loss}
\label{sup:style_loss}
\vspace{-0.5\baselineskip}

As described in Sec.~\ref{sec:method}~\refinpaper{}, any off-the-shelf perceptual similarity loss can be utilized in our method.
Here we show the results of our method with Style Loss~\cite{gatys2016styleloss} as the loss function $\C{L}$ in Eq.~\ref{eq:gradient_descent}~\refinpaper{}.
Fig.~\ref{fig:supp_style_loss} shows panorama images generated by MultiDiffusion~\cite{bartal2023multidiffusion}, and our method with LPIPS~\cite{zhang2018lpips} and Style Loss~\cite{gatys2016styleloss} as the perceptual similarity loss function, respectively.
To observe visible changes in the panorama outputs, we multiplied $10^6$ to the Style Loss
and set the gradient descent weight $w$ to $0.1$. Tab.~\ref{tab:supp_full_quantitative} (row 13) demonstrates that \SyncDiffusion{} with Style Loss achieves better coherence compared to MultiDiffusion as measured by Intra-LPIPS and Intra-Style-L, while showing a negative effect on the metrics regarding fidelity: Mean-GIQA, FID, and KID. Note that Intra-Style-L is significantly decreased as the guidance was provided with Style Loss.
The second row in Fig.~\ref{fig:supp_style_loss} shows that Style Loss can guide the joint diffusion processes to generate a globally coherent panorama image, as compared to the MultiDiffusion output in the first row.

{
\begin{figure*}[h!]
    \small
    \centering
    \renewcommand{\arraystretch}{0.0}
    \setlength{\tabcolsep}{0.0em}
    \setlength{\fboxrule}{0.0pt}
    \setlength{\fboxsep}{0pt}
    
    \begin{tabularx}{\textwidth}{>{\centering\arraybackslash}m{0.1\textwidth} >{\centering\arraybackslash}m{0.9\textwidth}}
    & \textit{``Natural landscape in anime style illustration''} \\
    \scriptsize{MD~\cite{bartal2023multidiffusion}} & 
    \includegraphics[width=0.895\textwidth]{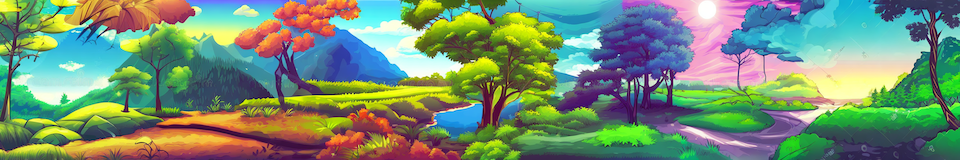}  \\
    \scriptsize{\makecell{\textsc{Sync-}\\\textsc{Diffusion}\\(Style Loss)}} & 
    \framebox{\includegraphics[width=0.895\textwidth]{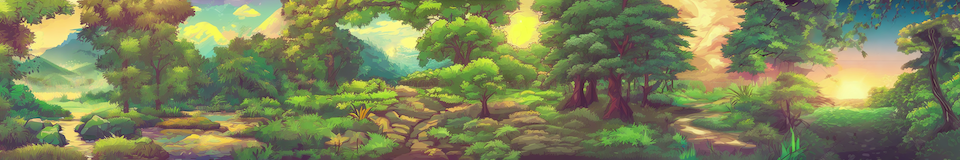}} \\
    \scriptsize{\makecell{\textsc{Sync-}\\\textsc{Diffusion}\\(LPIPS)}} & 
    \framebox{\includegraphics[width=0.895\textwidth]{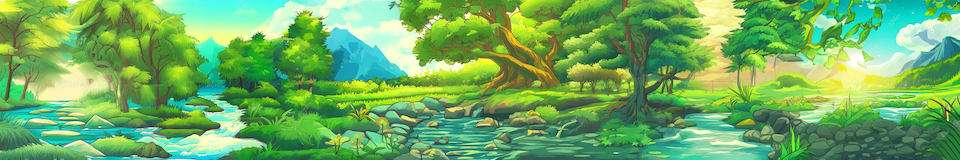}} \\
    \end{tabularx}
    \vspace{-0.5\baselineskip}
    \caption{Qualitative comparisons of MultiDiffusion and \SyncDiffusion{} with Style Loss.}
    \label{fig:supp_style_loss}
\end{figure*}
}

\input{sections/supp_ablation_x_t}
\input{sections/supp_sampling_time_analysis}
\input{sections/supp_userstudy}
\input{sections/supp_more_qualitative}

%% file: sections/supp_more_prompts.tex
\subsection{More Qualitative Results with Various Prompts}
More qualitative results with various prompts are shown in the figures below. The resolutions of images are $512 \times 3072$ for horizontal panoramas and $2048 \times 512$ for vertical panoramas.
\label{sup:more_prompts}

{
\small
\setlength{\tabcolsep}{0em}
\def\arraystretch{0.0}

\begin{longtable}[h!]{>{\centering\arraybackslash}m{0.10\textwidth}>{\centering\arraybackslash}m{0.10\textwidth}>{\centering\arraybackslash}m{0.10\textwidth}>{\centering\arraybackslash}m{0.10\textwidth}>{\centering\arraybackslash}m{0.10\textwidth}>{\centering\arraybackslash}m{0.10\textwidth}>{\centering\arraybackslash}m{0.10\textwidth}>{\centering\arraybackslash}m{0.10\textwidth}>{\centering\arraybackslash}m{0.10\textwidth}>{\centering\arraybackslash}m{0.10\textwidth}}
    \endhead
    & \multicolumn{9}{c}{\hspace*{0cm}\textit{``Skyline of New York City''}} \\
    \scriptsize{MD~\cite{bartal2023multidiffusion}}
    & \multicolumn{9}{c}{\hspace*{0cm}\raisebox{-0.5\totalheight}{\includegraphics[width=0.895\textwidth]{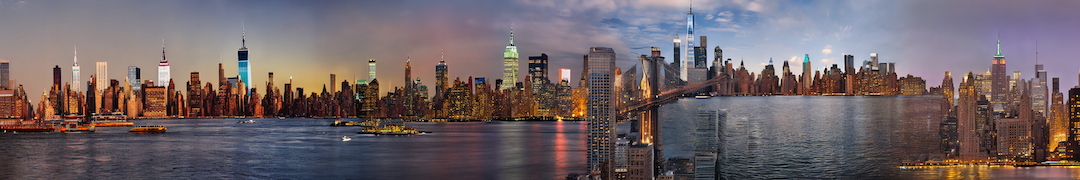}}}\\
    \\[+0.5em]
    \scriptsize{\makecell{\textsc{Sync-}\\\textsc{Diffusion}}} 
    & \multicolumn{9}{c}{\hspace*{0cm}\raisebox{-0.5\totalheight}{\includegraphics[width=0.895\textwidth]{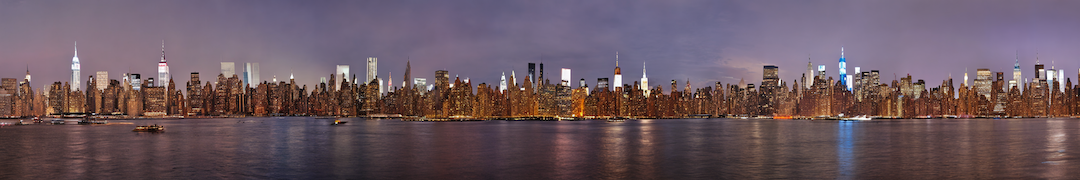}}}  \\
    & \multicolumn{9}{c}{\makecell{\\\hspace*{0cm}\textit{``An illustration of a beach in La La Land style''}}}  \\
    \scriptsize{MD~\cite{bartal2023multidiffusion}} \vspace{0.1cm}&
    \multicolumn{9}{c}{\hspace*{0cm}\raisebox{-0.5\totalheight}{\includegraphics[width=0.895\textwidth]{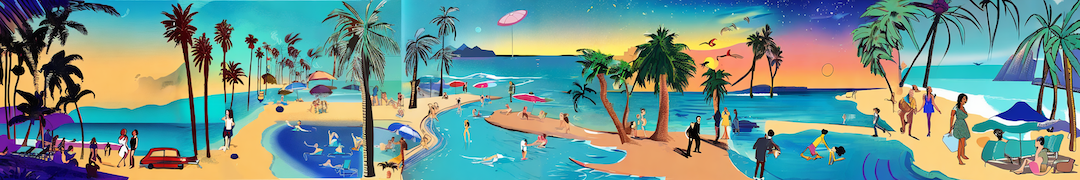}}}  \\
    \\[+0.5em]
    \scriptsize{\makecell{\textsc{Sync-}\\\textsc{Diffusion}}} &
    \multicolumn{9}{c}{\hspace*{0cm}\raisebox{-0.5\totalheight}{\includegraphics[width=0.895\textwidth]{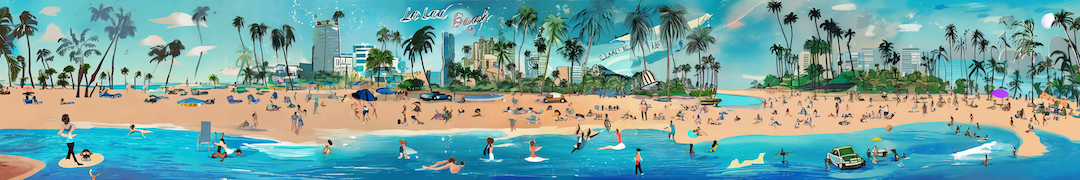}}}  \\
    \pagebreak
    
    \multicolumn{4}{c}{\makecell{\textit{``A waterfall''}\\}} & & &
    \multicolumn{4}{c}{\makecell{\textit{``A top view of a single railway''}\\}} \\
    \\[+0.5em]
    \multicolumn{2}{c}{\scriptsize{MD~\cite{bartal2023multidiffusion}}} & \multicolumn{2}{c}{\scriptsize{\textsc{SyncDiffusion}}} & & &
    \multicolumn{2}{c}{\scriptsize{MD~\cite{bartal2023multidiffusion}}} & \multicolumn{2}{c}{\scriptsize{\textsc{SyncDiffusion}}} \\

    \multicolumn{2}{c}{\includegraphics[width=0.20\textwidth]{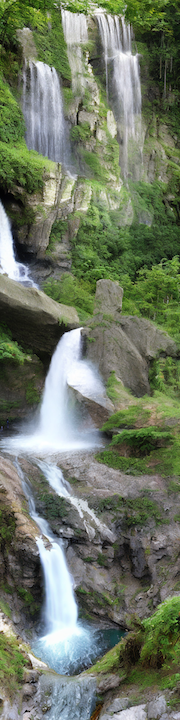}}\hspace{0.1cm}  & 
    \multicolumn{2}{c}{\includegraphics[width=0.20\textwidth]{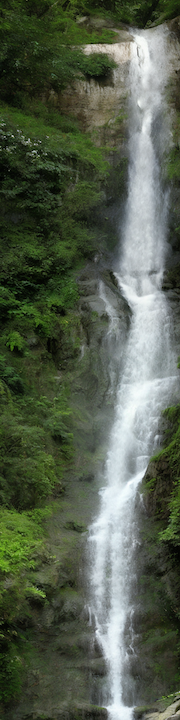}}  & & &
    \multicolumn{2}{c}{\includegraphics[width=0.20\textwidth]{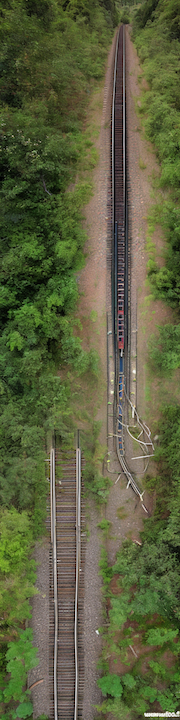}}\hspace{0.1cm}  & 
    \multicolumn{2}{c}{\includegraphics[width=0.20\textwidth]{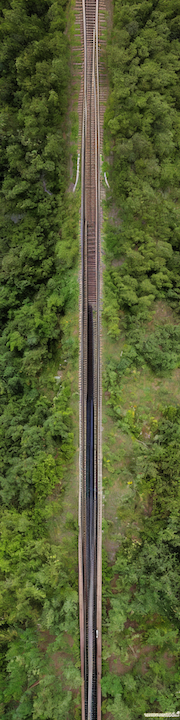}} \\
    \\[+1em]

    & \multicolumn{9}{c}{\makecell{\\\hspace*{0cm}\textit{``Silhouette wallpaper of a dreamy scene with shooting stars''}}}  \\
    \scriptsize{MD~\cite{bartal2023multidiffusion}} \vspace{0.1cm}&
    \multicolumn{9}{c}{\hspace*{0cm}\raisebox{-0.5\totalheight}{\includegraphics[width=0.895\textwidth]{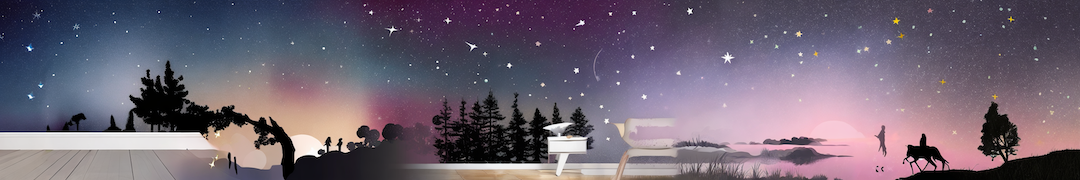}}}  \\
    \\[+0.5em]
    \scriptsize{\makecell{\textsc{Sync-}\\\textsc{Diffusion}}} &
    \multicolumn{9}{c}{\hspace*{0cm}\raisebox{-0.5\totalheight}{\includegraphics[width=0.895\textwidth]{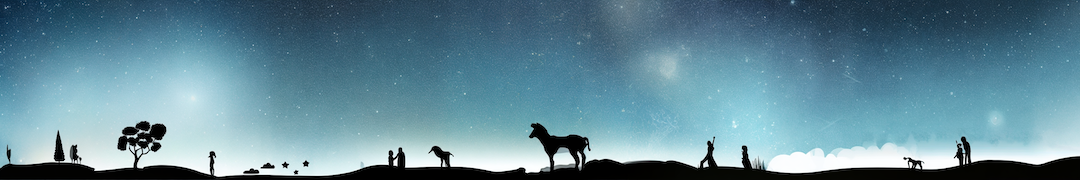}}}  \\
    & \multicolumn{9}{c}{\makecell{\makecell{\\\hspace*{0cm}\textit{``A photo of a rock concert''}}}}  \\
    \scriptsize{MD~\cite{bartal2023multidiffusion}} \vspace{0.1cm}&
    \multicolumn{9}{c}{\hspace*{0cm}\raisebox{-0.5\totalheight}{\includegraphics[width=0.895\textwidth]{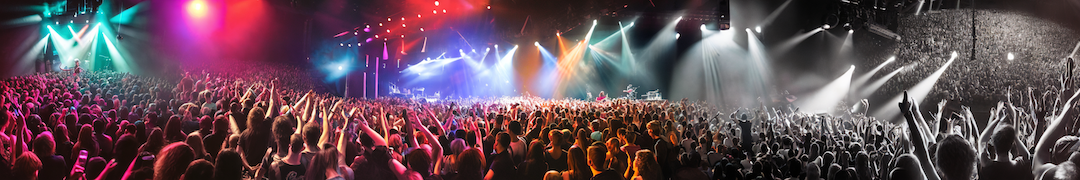}}}  \\
    \\[+0.5em]
    \scriptsize{\makecell{\textsc{Sync-}\\\textsc{Diffusion}}} &
    \multicolumn{9}{c}{\hspace*{0cm}\raisebox{-0.5\totalheight}{\includegraphics[width=0.895\textwidth]{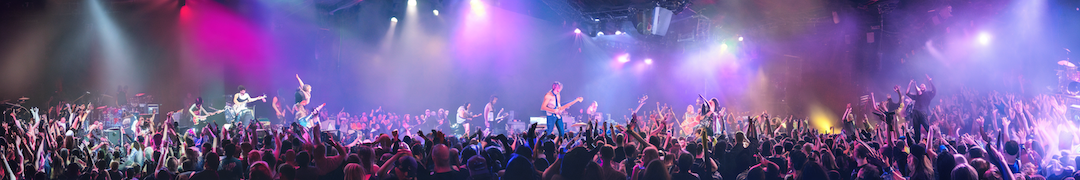}}}  \\
    \pagebreak

    \multicolumn{4}{c}{\makecell{\textit{``A photo of vines on a brick wall''}\\}} & & &
    \multicolumn{4}{c}{\makecell{\textit{``A bird’s eye view of an alley with shops''}\\}} \\
    \\[+0.5em]
    \multicolumn{2}{c}{\scriptsize{MD~\cite{bartal2023multidiffusion}}} & \multicolumn{2}{c}{\scriptsize{\textsc{SyncDiffusion}}} & & &
    \multicolumn{2}{c}{\scriptsize{MD~\cite{bartal2023multidiffusion}}} & \multicolumn{2}{c}{\scriptsize{\textsc{SyncDiffusion}}} \\

    \multicolumn{2}{c}{\includegraphics[width=0.20\textwidth]{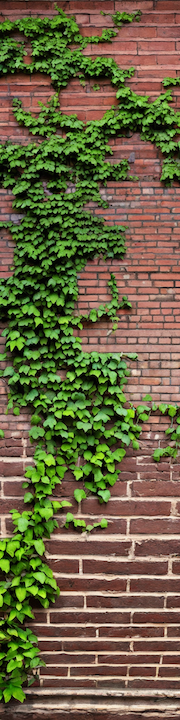}}  & 
    \multicolumn{2}{c}{\includegraphics[width=0.20\textwidth]{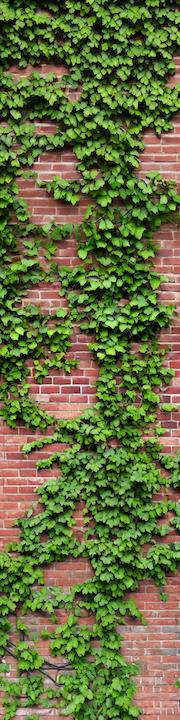}}  & & &
    \multicolumn{2}{c}{\includegraphics[width=0.20\textwidth]{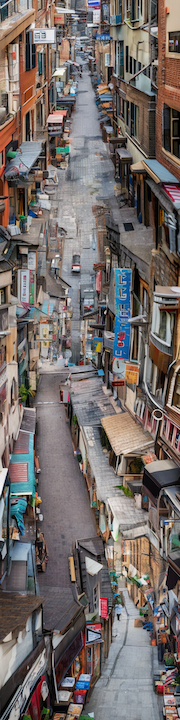}}  & 
    \multicolumn{2}{c}{\includegraphics[width=0.20\textwidth]{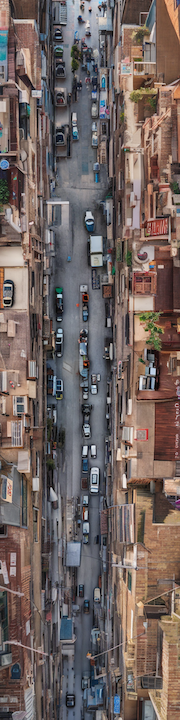}} \\
    \\[+1em]

    & \multicolumn{9}{c}{\makecell{\\ \hspace*{0cm}\textit{``A beach with palm trees''}}}  \\
    \scriptsize{MD~\cite{bartal2023multidiffusion}} \vspace{0.1cm}&
    \multicolumn{9}{c}{\hspace*{0cm}\raisebox{-0.5\totalheight}{\includegraphics[width=0.895\textwidth]{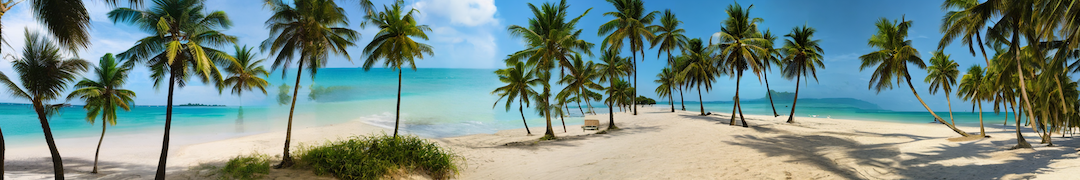}}}  \\
    \\[+0.5em]
    \scriptsize{\makecell{\textsc{Sync-}\\\textsc{Diffusion}}} &
    \multicolumn{9}{c}{\hspace*{0cm}\raisebox{-0.5\totalheight}{\includegraphics[width=0.895\textwidth]{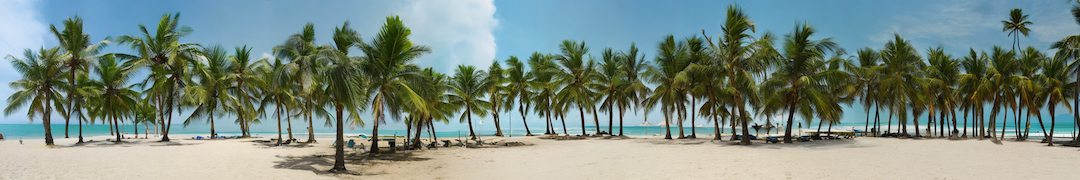}}}  \\
    & \multicolumn{9}{c}{\makecell{\\ \hspace*{0cm}\textit{``A photo of a grassland with animals''}}}  \\
    \scriptsize{MD~\cite{bartal2023multidiffusion}} \vspace{0.1cm}&
    \multicolumn{9}{c}{\hspace*{0cm}\raisebox{-0.5\totalheight}{\includegraphics[width=0.895\textwidth]{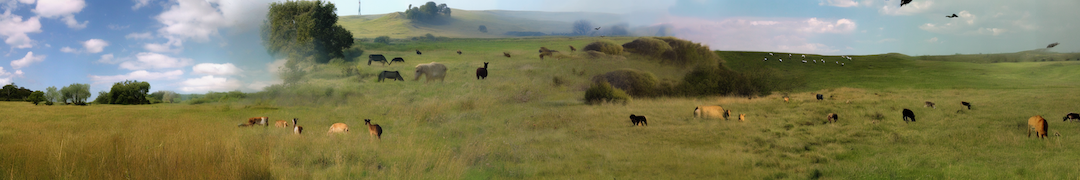}}}  \\
    \\[+0.5em]
    \scriptsize{\makecell{\textsc{Sync-}\\\textsc{Diffusion}}} &
    \multicolumn{9}{c}{\hspace*{0cm}\raisebox{-0.5\totalheight}{\includegraphics[width=0.895\textwidth]{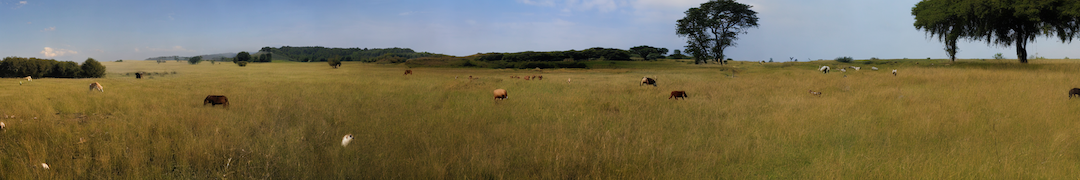}}}  \\
    \pagebreak
    & \multicolumn{9}{c}{\makecell{\\\hspace*{0cm}\textit{``A cinematic view of a castle in the sunset''}}}  \\
    \scriptsize{MD~\cite{bartal2023multidiffusion}} \vspace{0.1cm}&
    \multicolumn{9}{c}{\hspace*{0cm}\raisebox{-0.5\totalheight}{\includegraphics[width=0.895\textwidth]{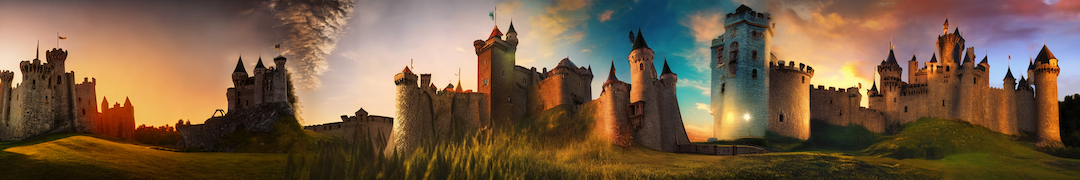}}}  \\
    \\[+0.5em]
    \scriptsize{\makecell{\textsc{Sync-}\\\textsc{Diffusion}}} &
    \multicolumn{9}{c}{\hspace*{0cm}\raisebox{-0.5\totalheight}{\includegraphics[width=0.895\textwidth]{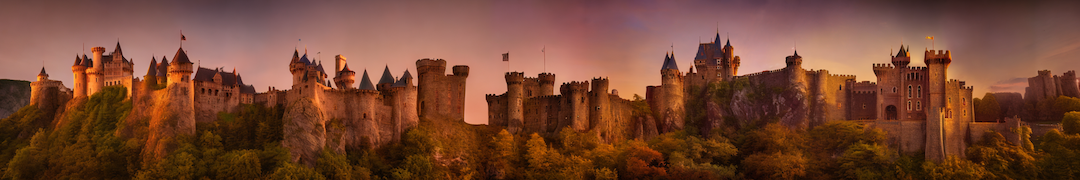}}}  \\
    & \multicolumn{9}{c}{\makecell{\\\hspace*{0cm} \textit{``A film photo of a beachside street under the sunset''}}}  \\
    \scriptsize{MD~\cite{bartal2023multidiffusion}} \vspace{0.1cm}&
    \multicolumn{9}{c}{\hspace*{0cm}\raisebox{-0.5\totalheight}{\includegraphics[width=0.895\textwidth]{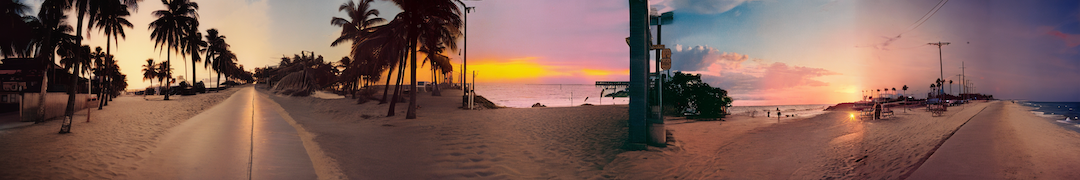}}}  \\
    \\[+0.5em]
    \scriptsize{\makecell{\textsc{Sync-}\\\textsc{Diffusion}}} &
    \multicolumn{9}{c}{\hspace*{0cm}\raisebox{-0.5\totalheight}{\includegraphics[width=0.895\textwidth]{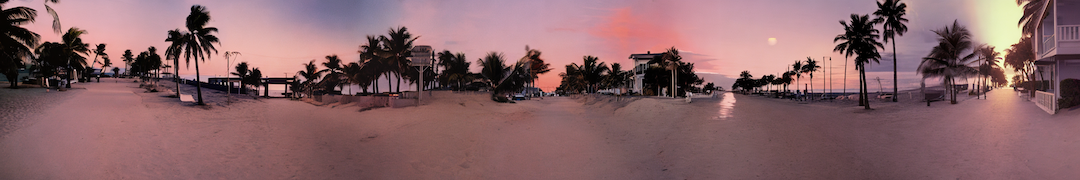}}}  \\
    & \multicolumn{9}{c}{\makecell{\\\hspace*{0cm} \textit{``A photo of a beautiful ocean with coral reef''}}}  \\
    \scriptsize{MD~\cite{bartal2023multidiffusion}} \vspace{0.1cm}&
    \multicolumn{9}{c}{\hspace*{0cm}\raisebox{-0.5\totalheight}{\includegraphics[width=0.895\textwidth]{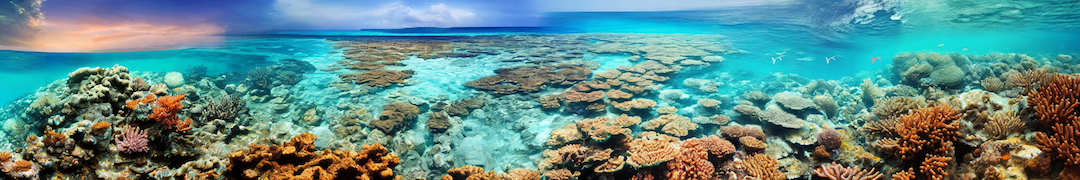}}}  \\
    \\[+0.5em]
    \scriptsize{\makecell{\textsc{Sync-}\\\textsc{Diffusion}}} &
    \multicolumn{9}{c}{\hspace*{0cm}\raisebox{-0.5\totalheight}{\includegraphics[width=0.895\textwidth]{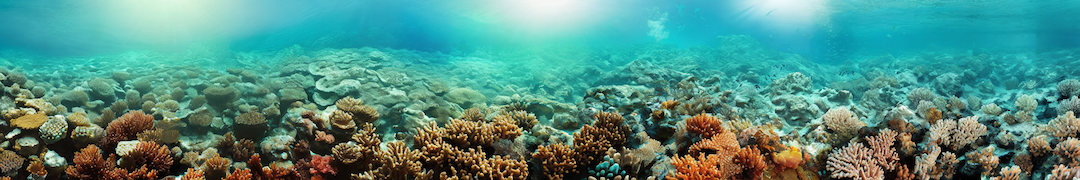}}}  \\
    & \multicolumn{9}{c}{\makecell{\\ \hspace*{0cm}\textit{``A photo of a lake under the northern lights''}}} \\
    \scriptsize{MD~\cite{bartal2023multidiffusion}} \vspace{0.1cm}&
    \multicolumn{9}{c}{\hspace*{0cm}\raisebox{-0.5\totalheight}{\includegraphics[width=0.895\textwidth]{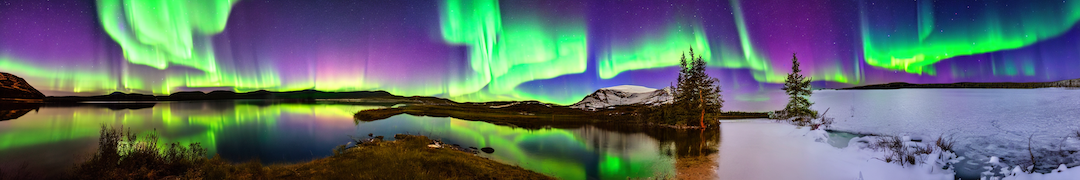}}}  \\
    \\[+0.5em]
    \scriptsize{\makecell{\textsc{Sync-}\\\textsc{Diffusion}}} &
    \multicolumn{9}{c}{\hspace*{0cm}\raisebox{-0.5\totalheight}{\includegraphics[width=0.895\textwidth]{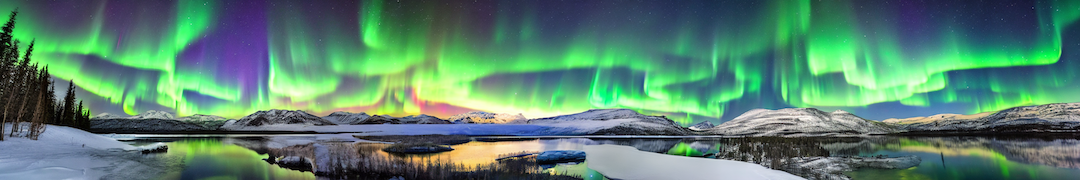}}}  \\
    
\end{longtable}
}

\renewcommand\tabularxcolumn[1]{m{#1}}
\newcolumntype{V}{>{\centering\arraybackslash}m{0.25\textwidth}}

%% file: sections/supp_more_quantitative.tex
\subsection{Details About Quantitative Evaluation}
\label{sup:more_quantitative}

\begin{figure}[h]
    \centering
    \includegraphics[width=0.95\textwidth]{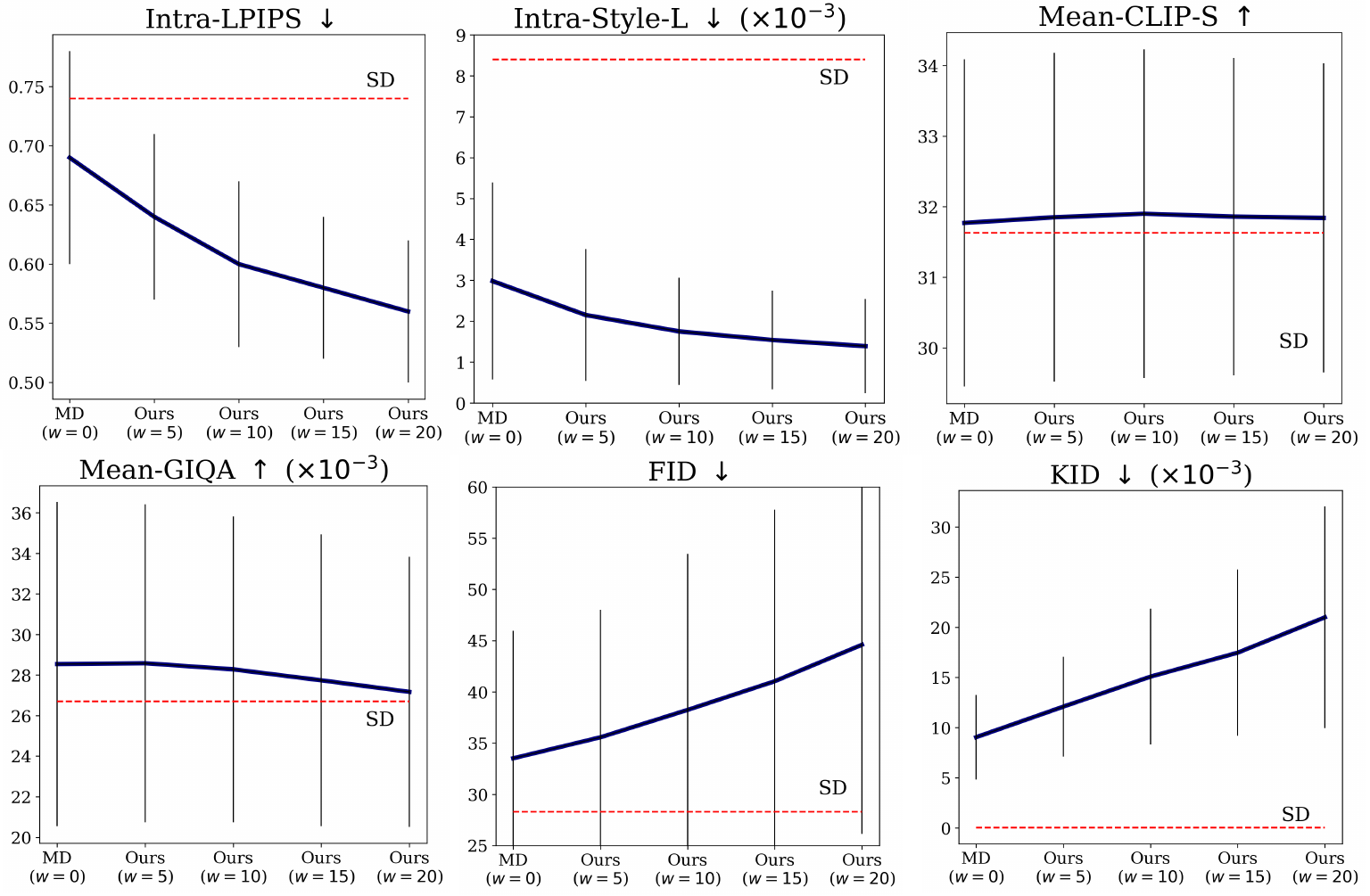}
    \caption{Line plots of the quantitative results shown in Tab.~\ref{tab:supp_full_quantitative} with varying gradient descent weight $w$. The dashed lines (SD) represent the evaluation results of the Stable Diffusion~\cite{rombach2022highresolution} reference set images. The vertical lines represent the standard deviation.}
    \label{fig:supp-graph}
    \vspace{-0.5\baselineskip}
\end{figure}

Tab.~\ref{tab:supp_full_quantitative} shows the detailed quantitative results of \textsc{SyncDiffusion} on panorama generation, reported in Fig.~\ref{fig:main-graph} \refofpaper{}.
Here we additionally show the results with the gradient descent weight $w=5$ and $w=15$, along with the weights $w=10$ and $w=20$ reported in Sec.~\ref{sec:results} \refofpaper{}. Note that we used KNN-GIQA~\cite{gu2020giqa} with $K=8$ to measure Mean-GIQA in all our experiments. As shown in Tab.~\ref{tab:supp_full_quantitative} (rows 3-7), as the gradient descent weight $w$ increases from $0$ to $20$, the results of our method display a significant improvement in global coherence, as shown in Intra-LPIPS~\cite{zhang2018lpips} which decreases from 0.69 ($w=0$) to 0.56 ($w=20$), and Intra-Style-L~\cite{gatys2016styleloss} which decreases from 2.98 ($w=0$) to 1.39 ($w=20$). These results are more apparent in the line plot of Intra-LPIPS and Intra-Style-L displayed in Fig.~\ref{fig:supp-graph}. Fig.~\ref{fig:weight_tradeoff} shows the qualitative comparison of the panorama images generated with different weights.

\setlength{\tabcolsep}{0em}
\def\arraystretch{0.0}
\renewcommand\tabularxcolumn[1]{m{#1}}
\newcolumntype{A}{>{\centering\arraybackslash}m{0.1\textwidth}}
\newcolumntype{B}{>{\centering\arraybackslash}m{0.9\textwidth}}

{
\begin{figure*}[h!]
    \small
    \centering
    \renewcommand{\arraystretch}{0.0}
    \setlength{\tabcolsep}{0.0em}
    \setlength{\fboxrule}{0.0pt}
    \setlength{\fboxsep}{0pt}
    
    \begin{tabularx}{\textwidth}{>{\centering\arraybackslash}m{0.1\textwidth} >{\centering\arraybackslash}m{0.9\textwidth}}
    & \textit{``A photo of a city skyline at night''} \\
    \scriptsize{\makecell{MD~\cite{bartal2023multidiffusion} \\ (\textsc{Sync-}\\\textsc{Diffusion} \\ $w=0$) }} & 
    \includegraphics[width=0.92\textwidth]{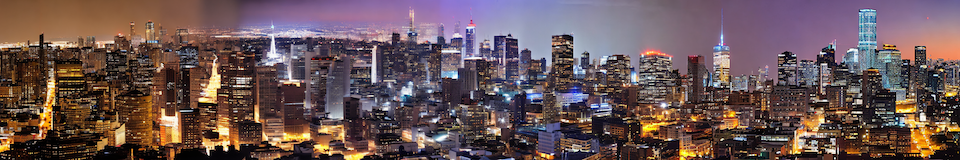}  \\
    \scriptsize{\makecell{\textsc{Sync-}\\\textsc{Diffusion}\\($w=5$)}} & 
    \includegraphics[width=0.92\textwidth]{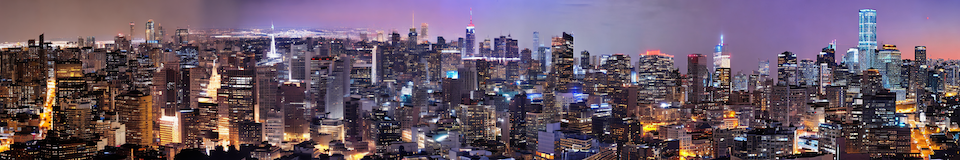} \\
    \scriptsize{\makecell{\textsc{Sync-}\\\textsc{Diffusion}\\($w=10$)}} & 
    \includegraphics[width=0.92\textwidth]{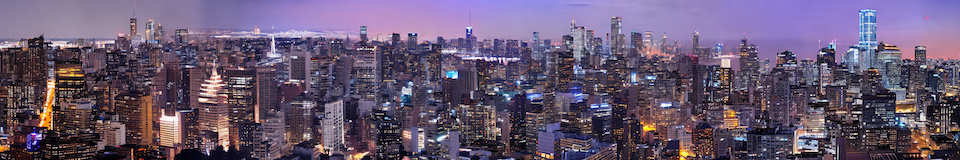} \\
    \scriptsize{\makecell{\textsc{Sync-}\\\textsc{Diffusion}\\($w=15$)}} & 
    \includegraphics[width=0.92\textwidth]{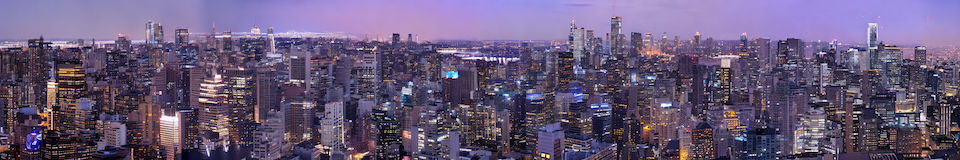}  \\
    \scriptsize{\makecell{\textsc{Sync-}\\\textsc{Diffusion}\\($w=20$)}} & 
    \includegraphics[width=0.92\textwidth]{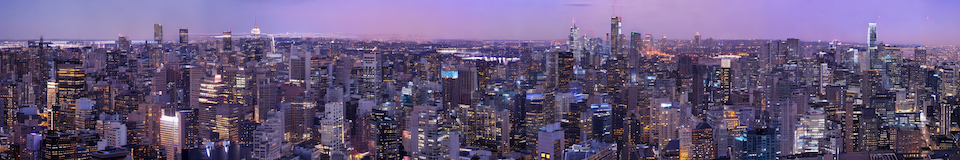} \\
    \end{tabularx}
    \caption{Qualitative comparison of different weights $w$. As $w$ increases, the generated panorama image gradually becomes globally coherent. Compared to MultiDiffusion, as $w$ increases, the left and right sides of the panorama image become more coherent.}
    \label{fig:weight_tradeoff}
\end{figure*}
}

\begin{table}[]
    \centering
    {\scriptsize
    \setlength{\tabcolsep}{0.2em}
    \begin{tabularx}{\linewidth}{>{\centering}m{0.3cm}>{\centering\arraybackslash}m{0.13\textwidth}>{\centering\arraybackslash}m{0.13\textwidth}>{\centering\arraybackslash}m{0.13\textwidth}>{\centering\arraybackslash}m{0.13\textwidth}>{\centering\arraybackslash}m{0.13\textwidth}>{\centering\arraybackslash}m{0.13\textwidth}>{\centering\arraybackslash}m{0.13\textwidth}}
    \toprule
    \multicolumn{2}{c}{} & \multicolumn{1}{c}{\multirow{2}{*}{Intra-LPIPS $\downarrow$}} & \multicolumn{1}{c}{Intra-Style-L $\downarrow$} & 
    \multicolumn{1}{c}{Mean-GIQA $\uparrow$} & \multicolumn{1}{c}{\multirow{2}{*}{FID $\downarrow$}} & \multicolumn{1}{c}{KID $\downarrow$} & \multicolumn{1}{c}{\multirow{2}{*}{Mean-CLIP-S $\uparrow$}} %
    \\[0.5em] & \multicolumn{2}{c}{} & \multicolumn{1}{c}{\tiny($\times 10^{-3}$)} & \multicolumn{1}{c}{\tiny($\times 10^{-3}$)} & \multicolumn{1}{c}{} & \multicolumn{1}{c}{\tiny($\times 10^{-3}$)} & \multicolumn{1}{c}{} \\
    \midrule
    1 & SD~\cite{rombach2022highresolution} & {\footnotesize 0.74} {\tiny $\pm$ 0.07} & {\footnotesize 8.40} {\tiny $\pm$ 6.27} & {\footnotesize 26.70} {\tiny $\pm$ 6.90} & {\footnotesize 28.31} {\tiny $\pm$ 10.89} & < {\footnotesize 0.01}{\tiny $\pm$ 0.13} & {\footnotesize 31.63} {\tiny $\pm$ 1.89} \\[1em]
    2 & BLD~\cite{avrahami2023blended_latent} & {\footnotesize 0.58} {\tiny $\pm$ 0.06} & {\footnotesize 4.64} {\tiny $\pm$ 3.32} & {\footnotesize 24.27} {\tiny $\pm$ 6.19} & {\footnotesize 84.29} {\tiny $\pm$ 36.74} & {\footnotesize 66.54} {\tiny $\pm$ 37.30} & {\footnotesize 31.41} {\tiny $\pm$ 1.66} \\[1em]
    \midrule
    \multicolumn{8}{c}{\SyncDiffusion{} with Various Gradient Descent Weight $w$ (Eq.~\ref{eq:gradient_descent})}\\[0.5em]
    \midrule
    3 & $w=0$ (MD~\cite{bartal2023multidiffusion}) & {\footnotesize 0.69} {\tiny $\pm$ 0.09} & {\footnotesize 2.98} {\tiny $\pm$ 2.41} & {\footnotesize 28.54} {\tiny $\pm$ 7.99} & \textbf{{\footnotesize 33.52} {\tiny $\pm$ 12.43}} & \textbf{{\footnotesize 9.04} {\tiny $\pm$ 4.23}} & {\footnotesize 31.77} {\tiny $\pm$ 2.32} \\[1em]
    
    4 & $w=5$ & {\footnotesize 0.64} {\tiny $\pm$ 0.07} & {\footnotesize 2.15} {\tiny $\pm$ 1.61} & \textbf{{\footnotesize 28.58} {\tiny $\pm$ 7.84}} & {\footnotesize 35.57} {\tiny $\pm$ 12.43} & {\footnotesize 12.09} {\tiny $\pm$ 4.98} & {\footnotesize 31.85} {\tiny $\pm$ 2.33} \\[1em]
    5 & $w=10$ & {\footnotesize 0.60} {\tiny $\pm$ 0.07} & {\footnotesize 1.75} {\tiny $\pm$ 1.31} & {\footnotesize 28.28} {\tiny $\pm$ 7.54} & {\footnotesize 38.24} {\tiny $\pm$ 15.24} & {\footnotesize 15.08} {\tiny $\pm$ 6.77} & \textbf{{\footnotesize 31.90} {\tiny $\pm$ 2.33}} \\[1em]
    6 & $w=15$ & {\footnotesize 0.58} {\tiny $\pm$ 0.06} & {\footnotesize 1.54} {\tiny $\pm$ 1.21} & {\footnotesize 27.74} {\tiny $\pm$ 7.19} & {\footnotesize 41.04} {\tiny $\pm$ 16.74} & {\footnotesize 17.47} {\tiny $\pm$ 8.29} & {\footnotesize 31.86} {\tiny $\pm$ 2.25} \\[1em]
    7 & $w=20$ & \textbf{{\footnotesize 0.56} {\tiny $\pm$ 0.06}} & {\footnotesize 1.39} {\tiny $\pm$ 1.15} & {\footnotesize 27.17} {\tiny $\pm$ 6.66} & {\footnotesize 44.60} {\tiny $\pm$ 18.45} & {\footnotesize 21.00} {\tiny $\pm$ 11.06} & {\footnotesize 31.84} {\tiny $\pm$ 2.19} \\[1em]
    
    \midrule
    \multicolumn{8}{c}{Panorama Size: $512\times2048$}\\[0.5em]
    \midrule
    8 & MD~\cite{bartal2023multidiffusion} & {\footnotesize 0.69} {\tiny $\pm$ 0.09} & {\footnotesize 2.96} {\tiny $\pm$ 2.41} & {\footnotesize 28.33} {\tiny $\pm$ 7.79} & {\footnotesize 33.07} {\tiny $\pm$ 12.38} & {\footnotesize 8.58} {\tiny $\pm$ 3.99} & {\footnotesize 31.77} {\tiny $\pm$ 2.14} \\[1em]
    9 & \SyncDiffusion{} & {\footnotesize 0.55} {\tiny $\pm$ 0.06} & {\footnotesize 1.39} {\tiny $\pm$ 1.19} & {\footnotesize 27.08} {\tiny $\pm$ 6.65} & {\footnotesize 43.33} {\tiny $\pm$ 17.98} & {\footnotesize 18.77} {\tiny $\pm$ 10.19} & {\footnotesize 31.77} {\tiny $\pm$ 2.14} \\[1em]

    \midrule
    \multicolumn{8}{c}{Panorama Size: $512\times1024$}\\[0.5em]
    \midrule
    10 & MD~\cite{bartal2023multidiffusion} & {\footnotesize 0.66} {\tiny $\pm$ 0.09} & {\footnotesize 2.57} {\tiny $\pm$ 1.97} & {\footnotesize 28.17} {\tiny $\pm$ 7.54} & {\footnotesize 30.66} {\tiny $\pm$ 11.79} & {\footnotesize 5.24} {\tiny $\pm$ 3.04} & {\footnotesize 31.73} {\tiny $\pm$ 2.22} \\[1em]
    11 & \SyncDiffusion{} & {\footnotesize 0.53} {\tiny $\pm$ 0.06} & {\footnotesize 1.09} {\tiny $\pm$ 0.77} & {\footnotesize 26.41} {\tiny $\pm$ 6.38} & {\footnotesize 40.35} {\tiny $\pm$ 16.43} & {\footnotesize 13.20} {\tiny $\pm$ 7.61} & {\footnotesize 31.71} {\tiny $\pm$ 2.01} \\[1em]

    \midrule
    \multicolumn{8}{c}{\SyncDiffusion{} Ablation Study}\\[0.5em]
    \midrule
    12 & Eq.~\ref{eq:gradient_descent} $\rightarrow$ Eq.~\ref{eq:gradient_descent_x_t} & {\footnotesize 0.68} {\tiny $\pm$ 0.09} & {\footnotesize 2.95} {\tiny $\pm$ 2.39} & {\footnotesize 28.53} {\tiny $\pm$ 7.99} & {\footnotesize 33.58} {\tiny $\pm$ 0.09} & {\footnotesize 9.15} {\tiny $\pm$ 4.25} & {\footnotesize 31.78} {\tiny $\pm$ 2.32} \\[1em]
    13 & Style Loss~\cite{gatys2016styleloss} & {\footnotesize 0.64} {\tiny $\pm$ 0.10} & \textbf{{\footnotesize 1.08} {\tiny $\pm$ 1.10}} & {\footnotesize 25.74} {\tiny $\pm$ 6.31} & {\footnotesize 73.05} {\tiny $\pm$ 37.56} & {\footnotesize 56.64} {\tiny $\pm$ 39.58} & {\footnotesize 31.15} {\tiny $\pm$ 2.32} \\[1em] 
    \bottomrule
    \end{tabularx}
    \vspace{0.5\baselineskip}
    \caption{Quantitative results on panorama generation.}
    \label{tab:supp_full_quantitative}
    }
\end{table}

\vspace{-\baselineskip}

%% file: sections/supp_diff_resolution.tex
\subsection{Quantitative Evaluation on Different Resolutions}
\vspace{-0.5\baselineskip}
\label{sup:diff_resolution}
We show the quantitative results on different resolutions in Tab.~\ref{tab:supp_full_quantitative} (row 10-13). In addition to the original $512 \times 3072$ resolution, Tab.~\ref{tab:supp_full_quantitative} shows the quantitative comparison of \textsc{SyncDiffusion} and MultiDiffusion~\cite{bartal2023multidiffusion} for smaller resolution panoramas ($512 \times 2048$ and $512 \times 1024$). In Fig.~\ref{tab:supp_full_quantitative}, when comparing the rows 10 and 11, 8 and 9, 3 and 7 respectively, the gap of Intra-LPIPS between our method and MultiDiffusion is preserved (0.13, 0.14, and 0.13, respectively), meaning that our method constantly produces more coherent panoramas than MultiDiffusion regardless of the resolution. The gap of Intra-Style-L between our method and MultiDiffusion even increases as the resolution increases (1.48, 1.57, and 1.59, respectively). On the other hand, the gap of FID and KID between the two methods also increases as the resolution increases: 9.69, 10.26, 11.08 for FID and 7.96, 10.19, 11.96 for KID.
We hypothesize that the increase in FID and KID of our method with longer panoramas is due to the tendency that for certain images it is more difficult to find other images that can be merged into a single coherent panorama. The above results indicate that while our method can guide the joint diffusion process to generate highly coherent images regardless of the resolution, generating longer panoramas that are globally coherent can lead to a decrease in the diversity of generations, thus resulting in a negative effect on FID and KID. \\

%% file: sections/supp_ablation_x_t.tex
\vspace{-0.5\baselineskip}
\subsection{Ablation on Predicting the Foreseen Denoised Observation}
\vspace{-0.5\baselineskip}
\label{sup:x_t_ablation}

Tab.~\ref{tab:supp_full_quantitative} (row 12) shows the quantitative comparison of the panorama generations using our method and after substituting the original Eq.~\ref{eq:gradient_descent} \refinpaper{} with Eq.~\ref{eq:gradient_descent_x_t} in which the noisy image $\V{x}^{(i)}_t$ is decoded instead of utilizing the foreseen denoised observation $\phi_\theta(\V{x}^{(i)}_t, t)$. Although Intra-LPIPS is still slightly reduced compared to MultiDiffusion when using Eq.~\ref{eq:gradient_descent_x_t}, the change is negligible compared to that of the original formulation Eq.~\ref{eq:gradient_descent}. This result is straightforward as measuring the perceptual loss between noisy images would not provide meaningful guidance to the diffusion process, whereas comparing the perceptual similarity of \textit{foreseen denoised} observations can give a meaningful guidance for global coherence.

%% file: sections/supp_sampling_time_analysis.tex
\vspace{-0.5\baselineskip}
\subsection{Analysis on the Computation Time}
\vspace{-0.5\baselineskip}
\label{sup:sampling_time_analysis}

As our~\SyncDiffusion{} module requires the gradient descent computation, it introduces additional computational overhead during the sampling process.
Since our method is based on the DDIM reverse process with 50 timesteps, the gradient descent is applied 50 times. 
Here we examine two methods to accelerate the generation process while still ensuring a notable improvement in coherence: applying \SyncDiffusion{} on a fixed interval and on the initial sampling steps.

\paragraph{Fixed interval}
We define $f$ as the frequency of the gradient descent during the DDIM reverse process of \SyncDiffusion{}, with the default value of $f=50$. Tab.~\ref{tab:supp_sampling_time_analysis} shows the quantitative results and the computation time when the gradient descent is performed 10 times ($f=10$) and 5 times ($f=5$) in total with uniform intervals, with the gradient descent weight fixed to $w=20$. Although applying the gradient descent for every step leads to the highest global coherence with Intra-LPIPS of 0.56 and Intra-Style-L of 1.39, in practice applying the gradient descent for 5 or 10 times can still achieve meaningful improvement in the coherence compared to MultiDiffusion as shown in rows 3-5 of Tab.~\ref{tab:supp_sampling_time_analysis}, while reducing the computation time compared to the $f=50$ case. Note that Intra-LPIPS decreases from 0.69 to 0.62 and Intra-Style-L decreases from 2.98 to 2.14 for $f=10$.

\vspace{-0.5\baselineskip}
\paragraph{Initial steps}
We further analyze the effectiveness of performing the gradient descent for the initial sampling steps.
Rows 6-7 of Tab.~\ref{tab:supp_sampling_time_analysis} show the quantitative results and the computation time when the gradient descent is applyed for the initial five and three steps out of the total 50 steps, respectively. The gradient descent weight is fixed to $w=20$. Comparing row 4 and row 7 shows that by computing the \SyncDiffusion{} function for just the initial three steps is analogous to computing it for ten times at regular intervals in terms of coherence (Intra-LPIPS, Intra-Style-L) and superior in terms of fidelity and diversity (FID and KID), while taking less than 70\% of the latter's computation time.
The qualitative comparisons of the early-stage synchronization are shown in Fig.~\ref{fig:initial_sampling}.

\begin{table*}[h!]
    \centering
    {\scriptsize
    \begin{tabular}{>{\centering\arraybackslash}m{0.09\textwidth}>{\centering\arraybackslash}m{0.13\textwidth}>{\centering\arraybackslash}m{0.13\textwidth}>{\centering\arraybackslash}m{0.13\textwidth}>{\centering\arraybackslash}m{0.13\textwidth}>{\centering\arraybackslash}m{0.13\textwidth}>{\centering\arraybackslash}m{0.13\textwidth}>{\centering\arraybackslash}m{0.13\textwidth}}
    \toprule
    \multicolumn{1}{c}{} & \multicolumn{1}{c}{\makecell{Intra-\\LPIPS $\downarrow$}} & \multicolumn{1}{c}{\makecell{Intra-\\Style-L $\downarrow$\\{\tiny($\times 10^{-3}$)}}} & \multicolumn{1}{c}{\makecell{Mean-\\GIQA $\uparrow$\\{\tiny($\times 10^{-3}$)}}} & \multicolumn{1}{c}{FID $\downarrow$} & \multicolumn{1}{c}{\makecell{KID $\downarrow$\\{\tiny($\times 10^{-3}$)}}} & \multicolumn{1}{c}{\makecell{Mean-\\CLIP-S $\uparrow$}} & \multicolumn{1}{c}{Time(s)}\\
    \midrule
    SD~\cite{rombach2022highresolution} & {\footnotesize 0.74}  {\tiny $\pm$ 0.07}& {\footnotesize 8.40} {\tiny $\pm$ 6.27}& {\footnotesize 26.70} {\tiny$\pm$ 6.90}& \textbf{{\footnotesize 28.31} {\tiny$\pm$ 10.89}}& \textbf{{\footnotesize <0.01} {\tiny$\pm$ 0.13}}& {\footnotesize 31.63} {\tiny$\pm$ 1.89}& -\\
    \\[0.5em]
    MD~\cite{bartal2023multidiffusion} & {\footnotesize 0.69} {\tiny $\pm$ 0.09}& {\footnotesize 2.98} {\tiny $\pm$ 2.41}& \textbf{{\footnotesize 28.54} {\tiny $\pm$ 7.99}}& {\footnotesize 33.52} {\tiny $\pm$ 12.43}& {\footnotesize 9.04} {\tiny $\pm$ 4.23}& {\footnotesize 31.77} {\tiny $\pm$ 2.32}& {\footnotesize 46.10} {\tiny $\pm$ 1.07}\\
    \\[0.5em]

    \midrule
    \multicolumn{8}{c}{\SyncDiffusion{}}\\[0.5em]
    \midrule

    $f=50$ & \textbf{{\footnotesize 0.56} {\tiny $\pm$ 0.06}}& \textbf{{\footnotesize 1.39} {\tiny $\pm$ 1.15}}& {\footnotesize 27.17} {\tiny $\pm$ 6.66}& {\footnotesize 44.60} {\tiny $\pm$ 18.45}& {\footnotesize 21.00} {\tiny $\pm$ 11.06}& {\footnotesize 31.84} {\tiny $\pm$ 2.19}& {\footnotesize 339.53} {\tiny $\pm$ 2.85}\\ \\[0.5em]
    $f=10$ & {\footnotesize 0.62} {\tiny $\pm$ 0.07}& {\footnotesize 2.14} {\tiny $\pm$ 1.72}& {\footnotesize 28.43} {\tiny $\pm$ 7.75}& {\footnotesize 36.22} {\tiny $\pm$ 14.03}& {\footnotesize 12.84} {\tiny $\pm$ 5.59}& \textbf{{\footnotesize 31.85} {\tiny $\pm$ 2.27}}& {\footnotesize 104.83} {\tiny $\pm$ 3.38}\\
    \\[0.5em]
    $f=5$ & {\footnotesize 0.64} {\tiny $\pm$ 0.07}& {\footnotesize 2.33} {\tiny $\pm$ 1.83}& {\footnotesize 28.44} {\tiny $\pm$ 7.85}& {\footnotesize 35.18} {\tiny $\pm$ 13.31}& {\footnotesize 11.43} {\tiny $\pm$ 4.68}& {\footnotesize 31.81} {\tiny $\pm$ 2.24}& {\footnotesize 81.17} {\tiny $\pm$ 0.53}\\
    \\[0.5em]
    Init. 5 Steps & {\footnotesize 0.61} {\tiny $\pm$ 0.06}& {\footnotesize 1.96} {\tiny $\pm$ 1.36}& {\footnotesize 28.21} {\tiny $\pm$ 7.48}& {\footnotesize 36.31} {\tiny $\pm$ 13.83}& {\footnotesize 12.09} {\tiny $\pm$ 4.76}& {\footnotesize 31.77} {\tiny $\pm$ 2.25}& {\footnotesize 79.12} {\tiny $\pm$ 1.72}\\
    \\[0.5em]
    Init. 3 Steps & {\footnotesize 0.62} {\tiny $\pm$ 0.06}& {\footnotesize 2.07} {\tiny $\pm$ 1.40}& {\footnotesize 28.43} {\tiny $\pm$ 8.19}& {\footnotesize 35.40} {\tiny $\pm$ 12.99}& {\footnotesize 11.15} {\tiny $\pm$ 3.76}& {\footnotesize 31.79} {\tiny $\pm$ 2.26}& {\footnotesize 71.56} {\tiny $\pm$ 2.64}\\
    \\
    \bottomrule
    \end{tabular}
    \caption{Analysis on the computation time of our~\SyncDiffusion{} and MultiDiffusion \cite{bartal2023multidiffusion}.}
    \label{tab:supp_sampling_time_analysis}
    }
\end{table*}

%% file: sections/supp_userstudy.tex
\vspace{-0.5\baselineskip}
\subsection{Details of User Study}
\vspace{-0.5\baselineskip}
\label{sup:user_study}

For each user study, the order of the images was shuffled.
Given a total of 200 questions with a random pair of panoramas, we collected 20 responses each from the participants on Amazon Mechanical Turk who passed our five vigilance tasks.
The vigilance tasks were designed to distinguish our outputs from concatenations of Stable Diffusion images generated without joint diffusion.
Out of the 100 participants, 86, 90, 84 participants successfully completed all the vigilance tasks for the user study for coherence, image quality and prompt compatibility, respectively.

\begin{figure*}[h!]
    \centering
    \footnotesize{
        \renewcommand{\arraystretch}{0.0}
        \setlength{\tabcolsep}{0.0em}
        \setlength{\fboxrule}{0.0pt}
        \setlength{\fboxsep}{0pt}
        \begin{tabularx}{\textwidth}{>{\centering\arraybackslash}m{0.5\textwidth} >{\centering\arraybackslash}m{0.5\textwidth}}
        \includegraphics[width=0.49\textwidth]{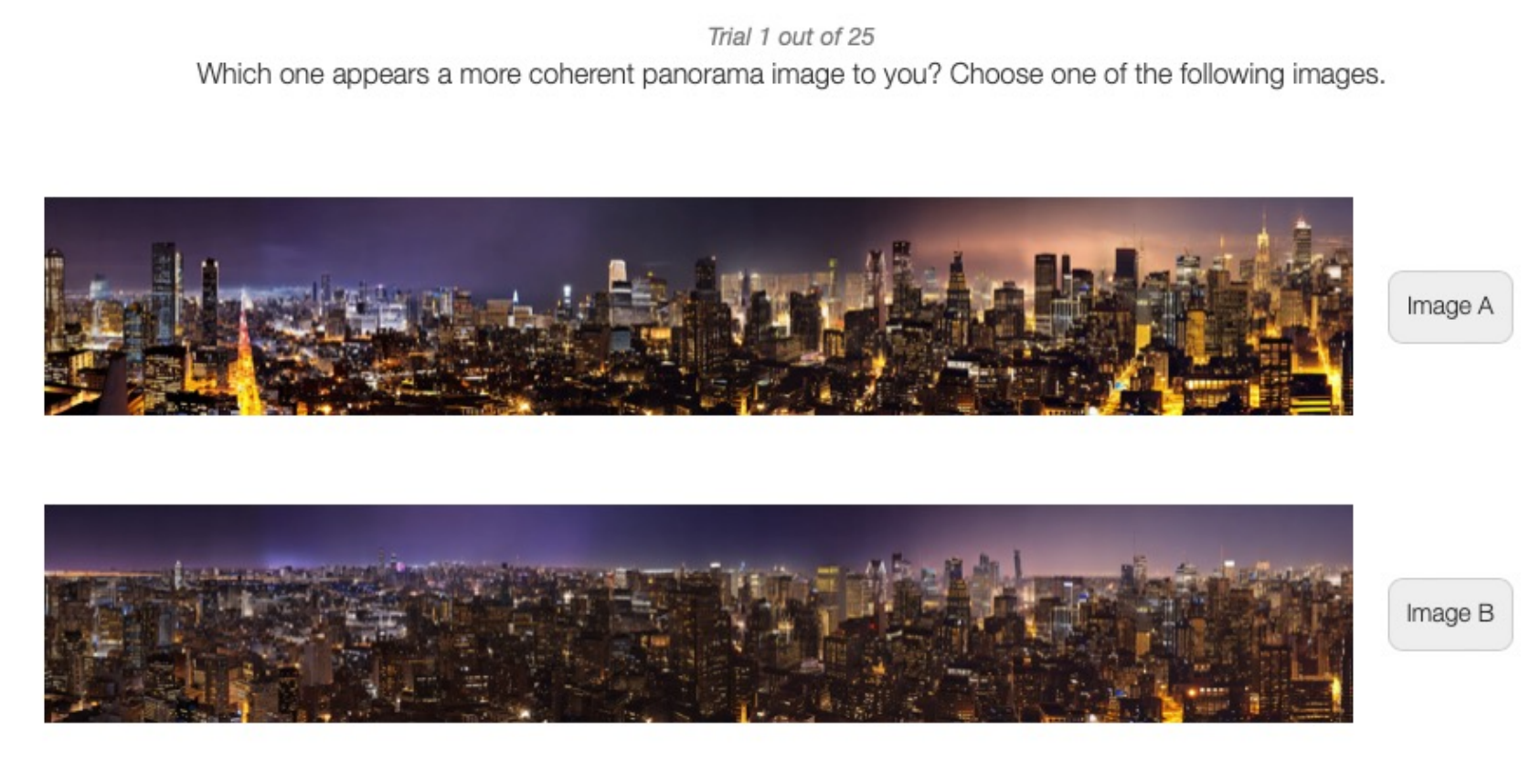} &
        \includegraphics[width=0.49\textwidth]{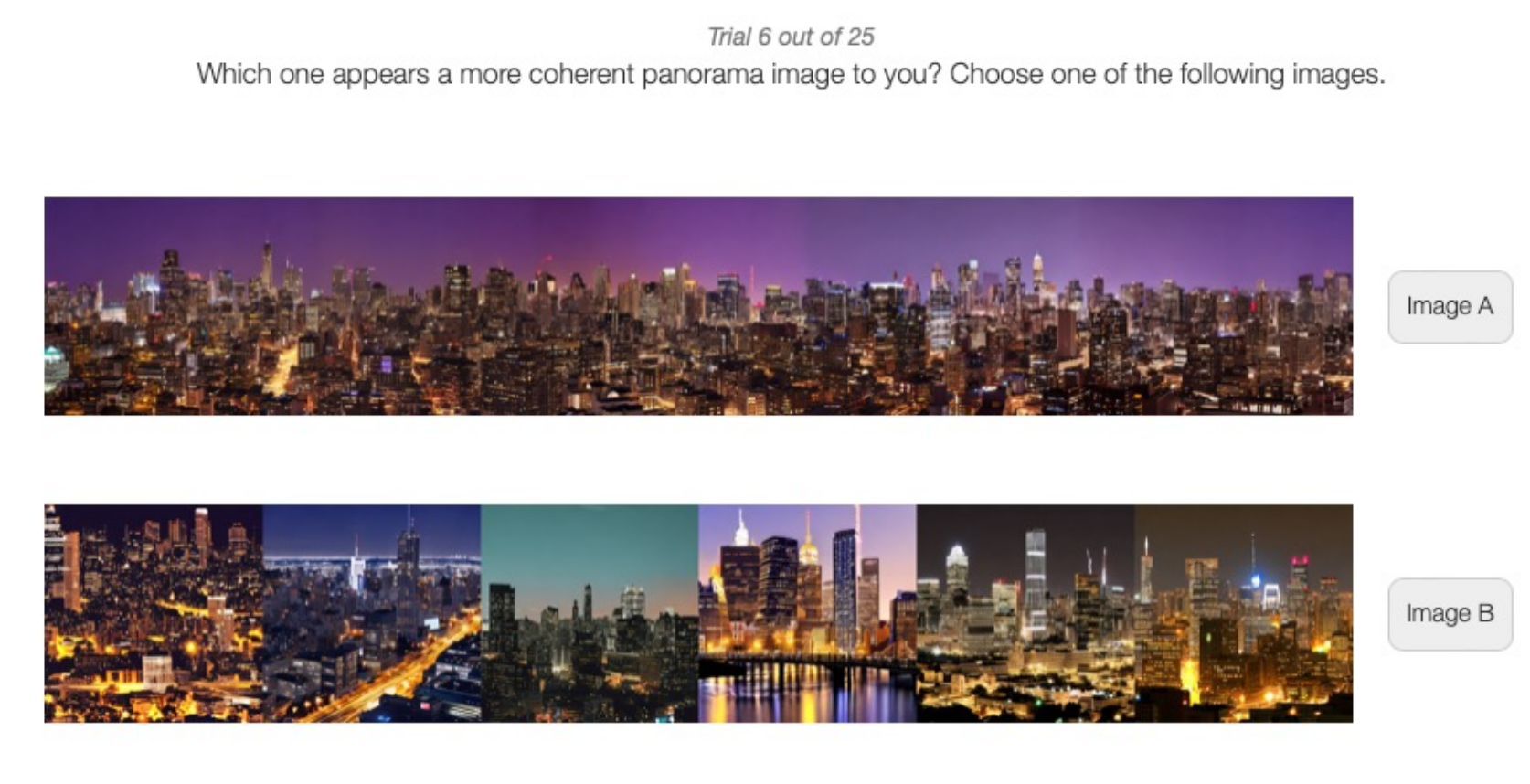}
        \end{tabularx}
    }
    \vspace{-0.75\baselineskip}
    \caption{User study screenshots.}
    \label{fig:userstudy}
\end{figure*}

Fig.~\ref{fig:userstudy} shows screenshots of our user study. We set all participants to be Amazon Mechanical Turk Masters who are located in the US. The average time that participants spent on solving a set of 25 problems (including the vigilance tasks) was 248.21 seconds, and we compensated them with a payment of $0.76\$$ per person. This is equal to $11.02\$$ per hour, which exceeds the US federal minimum wage.

\clearpage

{
\begin{figure*}[h!]
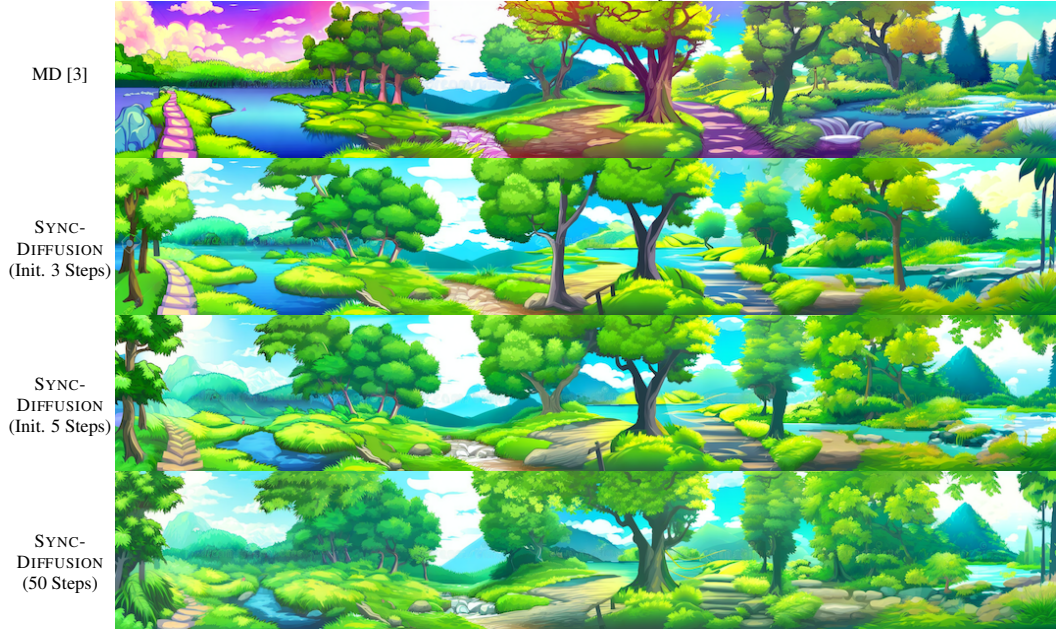

    \small
    \centering
    \renewcommand{\arraystretch}{0.0}
    \setlength{\tabcolsep}{0.0em}
    \setlength{\fboxrule}{0.0pt}
    \setlength{\fboxsep}{0pt}
    
    \begin{tabularx}{\textwidth}{>{\centering\arraybackslash}m{0.1\textwidth} >{\centering\arraybackslash}m{0.9\textwidth}}
    & \textit{``Natural landscape in anime style illustration''} \\
    \scriptsize{MD~\cite{bartal2023multidiffusion}} & 
    \includegraphics[width=0.895\textwidth]{figures/supp_resized/initial_steps_160/MD_21.png}  \\
    \scriptsize{\makecell{\textsc{Sync-}\\\textsc{Diffusion}\\(Init. 3 Steps)}} & 
    \framebox{\includegraphics[width=0.895\textwidth]{figures/supp_resized/initial_steps_160/Ours_3step_21.png}} \\
    \scriptsize{\makecell{\textsc{Sync-}\\\textsc{Diffusion}\\(Init. 5 Steps)}} & 
    \framebox{\includegraphics[width=0.895\textwidth]{figures/supp_resized/initial_steps_160/Ours_5step_21.png}} \\
    \scriptsize{\makecell{\textsc{Sync-}\\\textsc{Diffusion}\\(50 Steps)}} & 
    \framebox{\includegraphics[width=0.895\textwidth]{figures/supp_resized/initial_steps_160/Ours_full_21.png}} \\
    \end{tabularx}
    \vspace{-0.5\baselineskip}
    \caption{Qualitative comparisons of the early-stage synchronization of \SyncDiffusion{}.}
    \label{fig:initial_sampling}
\end{figure*}
}

%% file: sections/supp_more_qualitative.tex
\clearpage
\subsection{More Qualitative Results}
\label{sup:more_qualitative}

More qualitative results are shown in the figures below.

\setlength{\tabcolsep}{0em}
\def\arraystretch{0.0}
\renewcommand\tabularxcolumn[1]{m{#1}}
\newcolumntype{A}{>{\centering\arraybackslash}m{0.1\textwidth}}
\newcolumntype{B}{>{\centering\arraybackslash}m{0.9\textwidth}}

\begin{longtable}[h!]{AB}   
    \endhead
    \toprule
    \multicolumn{2}{c}{\textit{``A photo of a city skyline at night''}}  \\
    \midrule
    \scriptsize{BLD~\cite{avrahami2023blended_latent}} & 
    \includegraphics[width=0.9\textwidth]{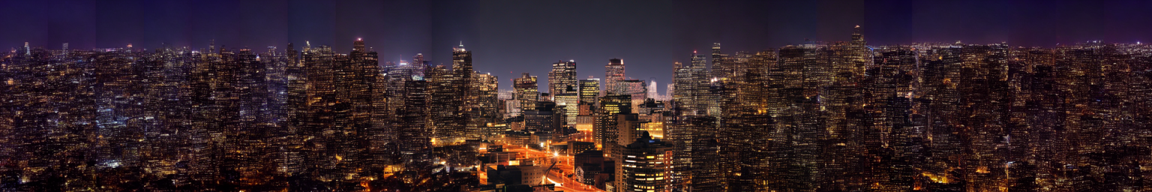} \\
    \scriptsize{MD~\cite{bartal2023multidiffusion}} & 
    \includegraphics[width=0.9\textwidth]{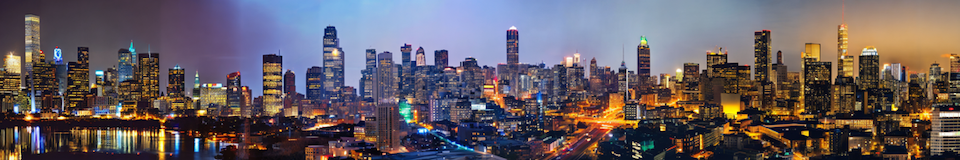}  \\
    \scriptsize{\makecell{Sync\\Diffusion}} & 
    \includegraphics[width=0.9\textwidth]{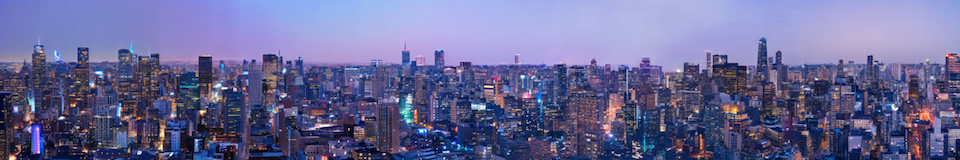} \\
    \midrule
    \scriptsize{BLD~\cite{avrahami2023blended_latent}} & 
    \includegraphics[width=0.9\textwidth]{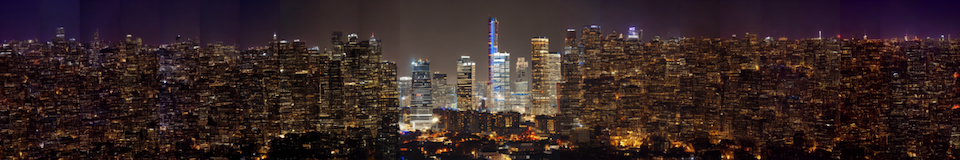} \\
    \scriptsize{MD~\cite{bartal2023multidiffusion}} & 
    \includegraphics[width=0.9\textwidth]{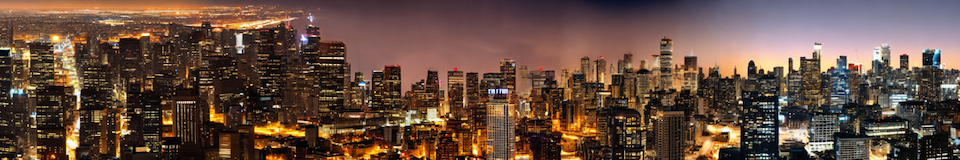}  \\
    \scriptsize{\makecell{Sync\\Diffusion}} & 
    \includegraphics[width=0.9\textwidth]{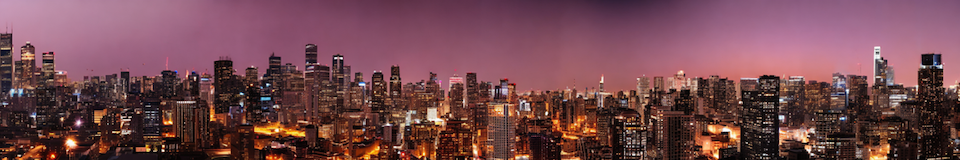} \\

    \midrule
    \scriptsize{BLD~\cite{avrahami2023blended_latent}} & 
    \includegraphics[width=0.9\textwidth]{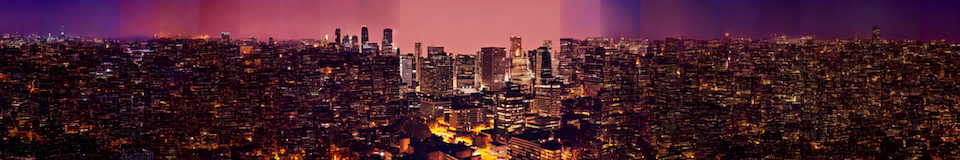} \\
    \scriptsize{MD~\cite{bartal2023multidiffusion}} & 
    \includegraphics[width=0.9\textwidth]{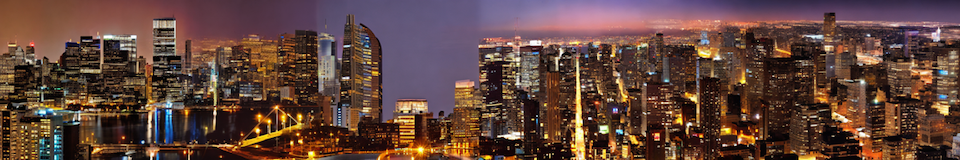}  \\
    \scriptsize{\makecell{Sync\\Diffusion}} & 
    \includegraphics[width=0.9\textwidth]{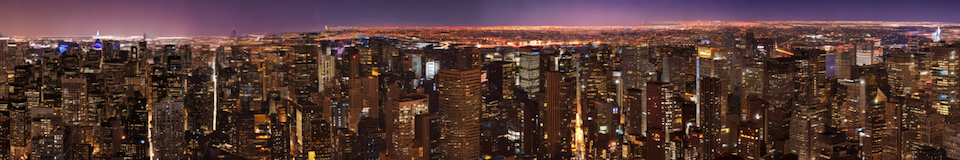} \\
    \midrule

    \pagebreak

    \toprule
    \multicolumn{2}{c}{\textit{``Natural landscape in anime style illustration''}} \\
    \midrule
    \scriptsize{BLD~\cite{avrahami2023blended_latent}} & 
    \includegraphics[width=0.9\textwidth]{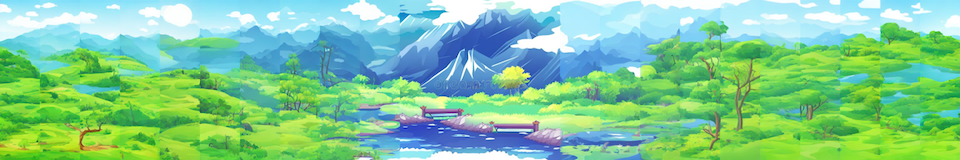} \\
    \scriptsize{MD~\cite{bartal2023multidiffusion}} & 
    \includegraphics[width=0.9\textwidth]{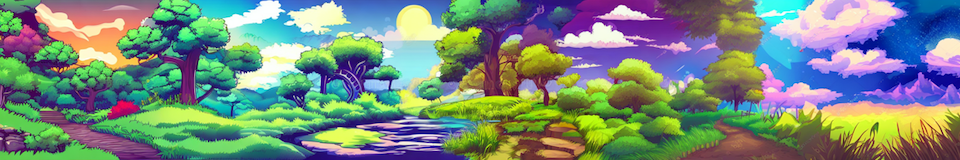}  \\
    \scriptsize{\makecell{Sync\\Diffusion}} & 
    \includegraphics[width=0.9\textwidth]{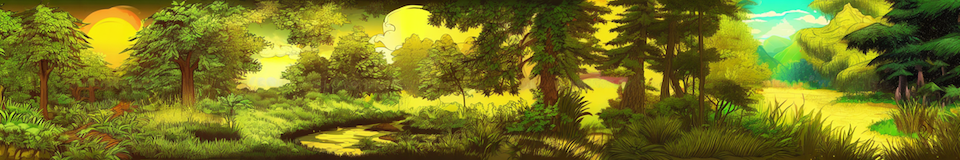} \\
    \midrule

    \scriptsize{BLD~\cite{avrahami2023blended_latent}} & 
    \includegraphics[width=0.9\textwidth]{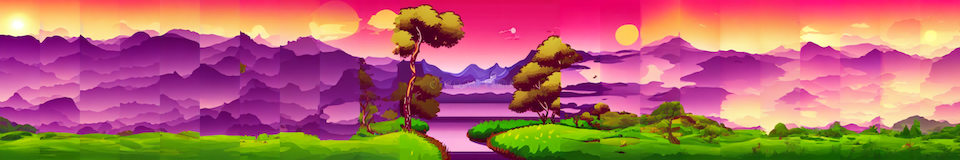} \\
    \scriptsize{MD~\cite{bartal2023multidiffusion}} & 
    \includegraphics[width=0.9\textwidth]{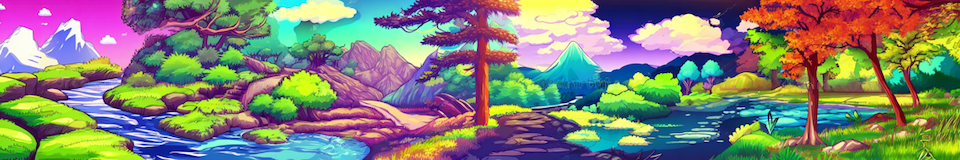}  \\
    \scriptsize{\makecell{Sync\\Diffusion}} & 
    \includegraphics[width=0.9\textwidth]{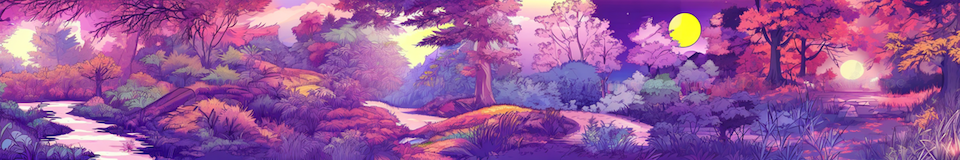} \\

    \midrule
    
    \scriptsize{BLD~\cite{avrahami2023blended_latent}} & 
    \includegraphics[width=0.9\textwidth]{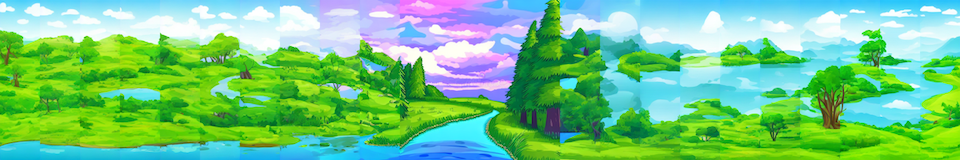} \\
    \scriptsize{MD~\cite{bartal2023multidiffusion}} & 
    \includegraphics[width=0.9\textwidth]{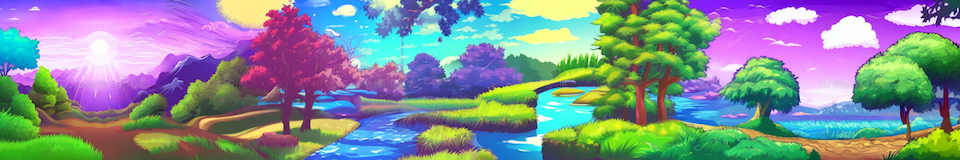}  \\
    \scriptsize{\makecell{Sync\\Diffusion}} & 
    \includegraphics[width=0.9\textwidth]{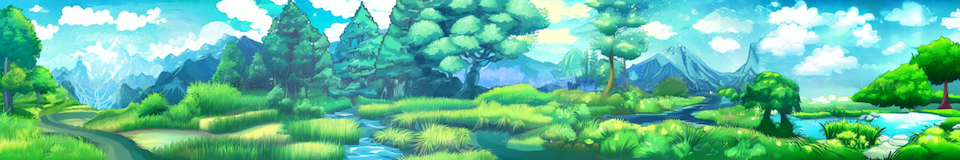} \\

    \bottomrule
    \pagebreak

    \toprule
    \multicolumn{2}{c}{\textit{``Natural landscape in anime style illustration''}} \\
    \midrule

    \scriptsize{BLD~\cite{avrahami2023blended_latent}} & 
    \includegraphics[width=0.9\textwidth]{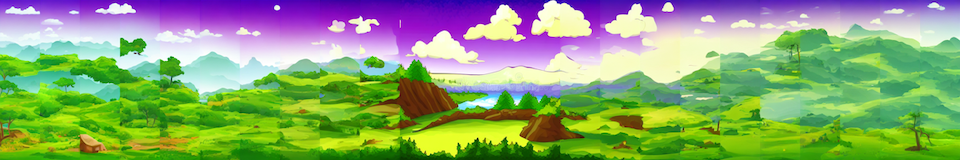} \\
    \scriptsize{MD~\cite{bartal2023multidiffusion}} & 
    \includegraphics[width=0.9\textwidth]{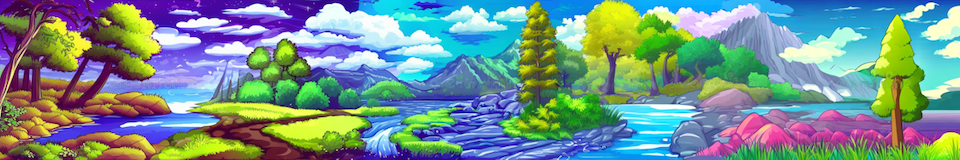}  \\
    \scriptsize{\makecell{Sync\\Diffusion}} & 
    \includegraphics[width=0.9\textwidth]{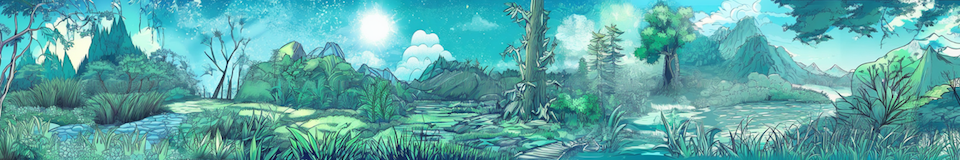} \\
    \bottomrule
    \multicolumn{2}{c}{\bigskip} \\

    \toprule
    \multicolumn{2}{c}{\textit{``A photo of a forest with a misty fog''}} \\
    \midrule
    \scriptsize{BLD~\cite{avrahami2023blended_latent}} & 
    \includegraphics[width=0.9\textwidth]{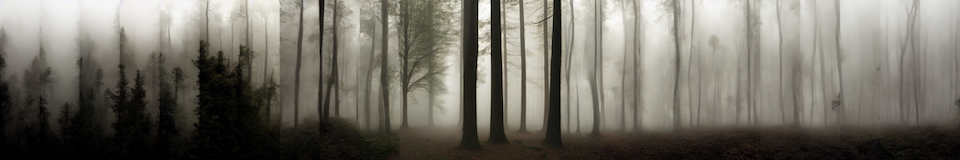} \\
    \scriptsize{MD~\cite{bartal2023multidiffusion}} & 
    \includegraphics[width=0.9\textwidth]{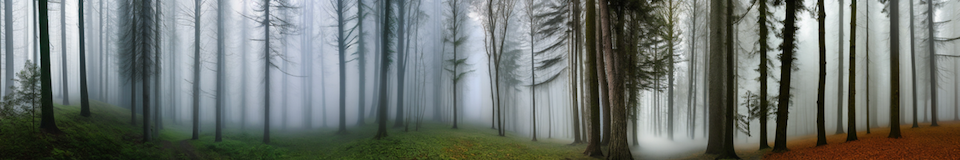}  \\
    \scriptsize{\makecell{Sync\\Diffusion}} & 
    \includegraphics[width=0.9\textwidth]{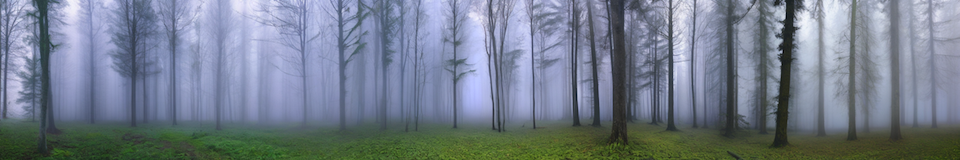} \\

    \midrule
    \scriptsize{BLD~\cite{avrahami2023blended_latent}} & 
    \includegraphics[width=0.9\textwidth]{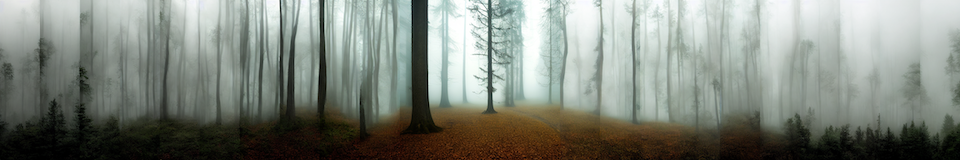} \\
    \scriptsize{MD~\cite{bartal2023multidiffusion}} & 
    \includegraphics[width=0.9\textwidth]{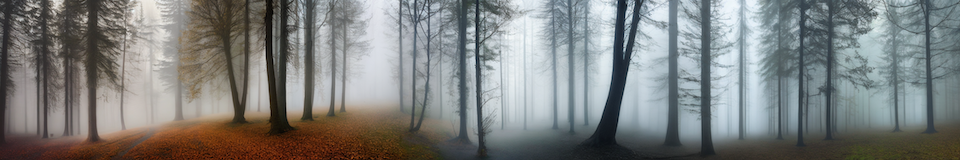}  \\
    \scriptsize{\makecell{Sync\\Diffusion}} & 
    \includegraphics[width=0.9\textwidth]{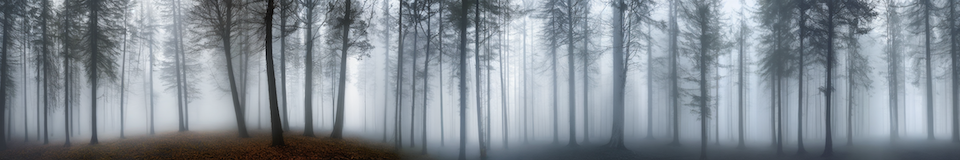} \\

    \bottomrule
    \pagebreak

    \toprule
    \multicolumn{2}{c}{\textit{``A photo of a snowy mountain peak with skiers''}} \\
    \midrule
    \scriptsize{BLD~\cite{avrahami2023blended_latent}} & 
    \includegraphics[width=0.9\textwidth]{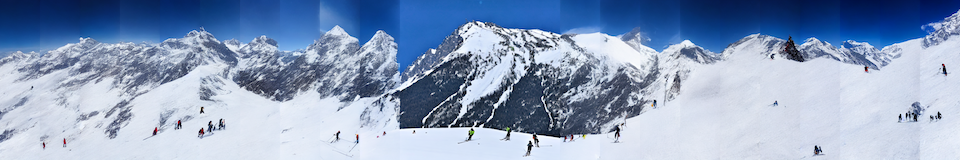} \\
    \scriptsize{MD~\cite{bartal2023multidiffusion}} & 
    \includegraphics[width=0.9\textwidth]{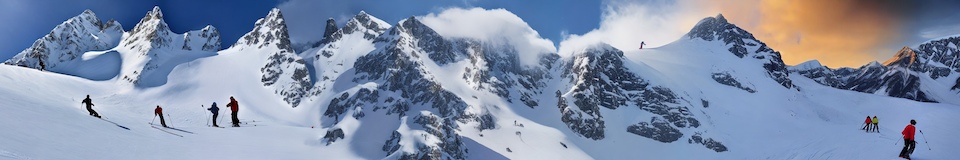}  \\
    \scriptsize{\makecell{Sync\\Diffusion}} & 
    \includegraphics[width=0.9\textwidth]{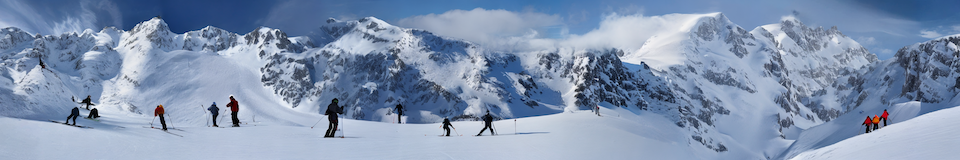} \\

    \bottomrule
    \multicolumn{2}{c}{\bigskip} \\

    \toprule
    \multicolumn{2}{c}{\textit{``A photo of a mountain range at twilight''}} \\
    \midrule
    \scriptsize{BLD~\cite{avrahami2023blended_latent}} & 
    \includegraphics[width=0.9\textwidth]{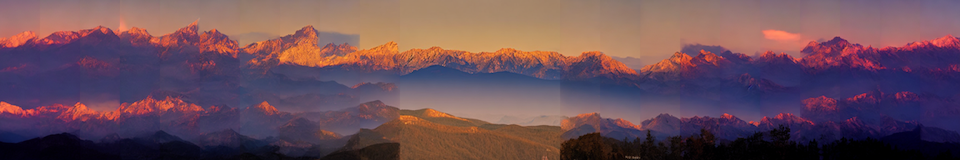} \\
    \scriptsize{MD~\cite{bartal2023multidiffusion}} & 
    \includegraphics[width=0.9\textwidth]{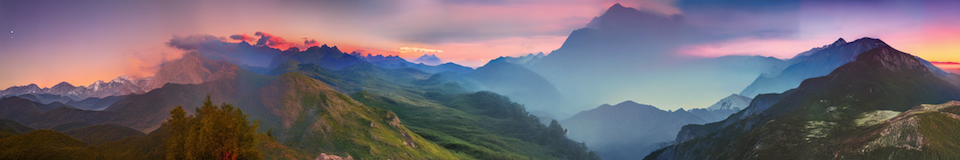}  \\
    \scriptsize{\makecell{Sync\\Diffusion}} & 
    \includegraphics[width=0.9\textwidth]{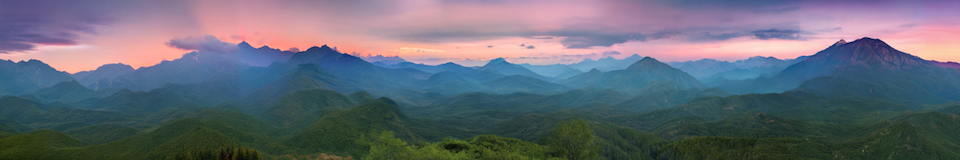} \\
    \midrule
    
    \scriptsize{BLD~\cite{avrahami2023blended_latent}} & 
    \includegraphics[width=0.9\textwidth]{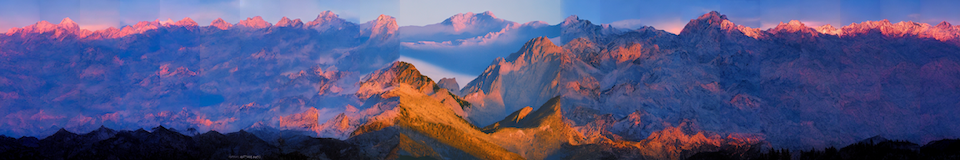} \\
    \scriptsize{MD~\cite{bartal2023multidiffusion}} & 
    \includegraphics[width=0.9\textwidth]{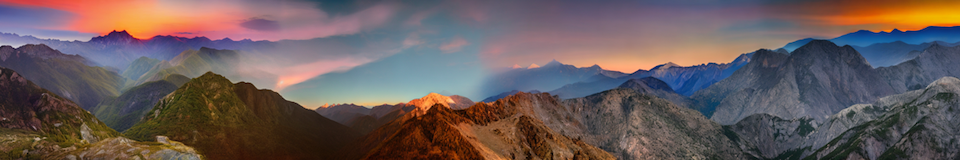}  \\
    \scriptsize{\makecell{Sync\\Diffusion}} & 
    \includegraphics[width=0.9\textwidth]{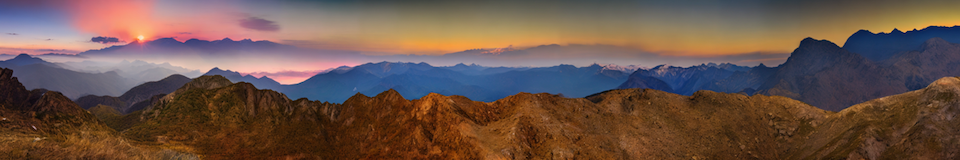} \\
    \bottomrule

    \pagebreak

    \toprule
    \multicolumn{2}{c}{\textit{``A photo of a mountain range at twilight''}} \\
    \midrule
    \scriptsize{BLD~\cite{avrahami2023blended_latent}} & 
    \includegraphics[width=0.9\textwidth]{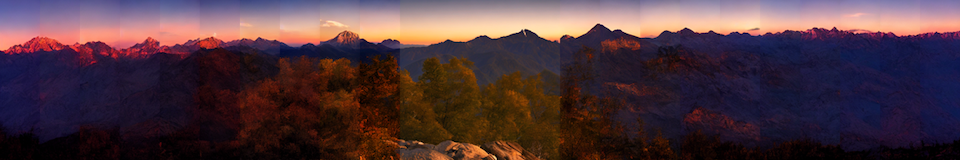} \\
    \scriptsize{MD~\cite{bartal2023multidiffusion}} & 
    \includegraphics[width=0.9\textwidth]{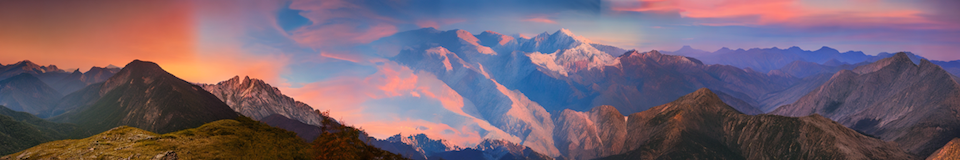}  \\
    \scriptsize{\makecell{Sync\\Diffusion}} & 
    \includegraphics[width=0.9\textwidth]{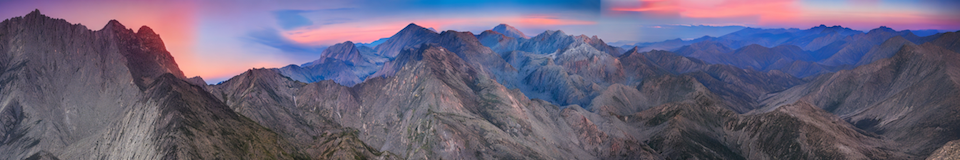} \\
    \midrule

    \scriptsize{BLD~\cite{avrahami2023blended_latent}} & 
    \includegraphics[width=0.9\textwidth]{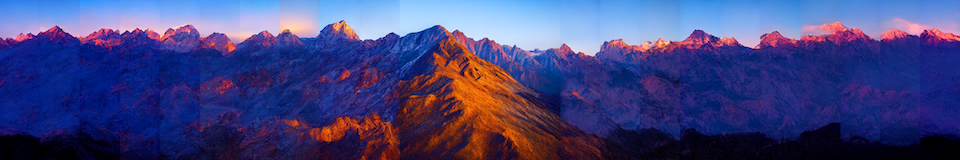} \\
    \scriptsize{MD~\cite{bartal2023multidiffusion}} & 
    \includegraphics[width=0.9\textwidth]{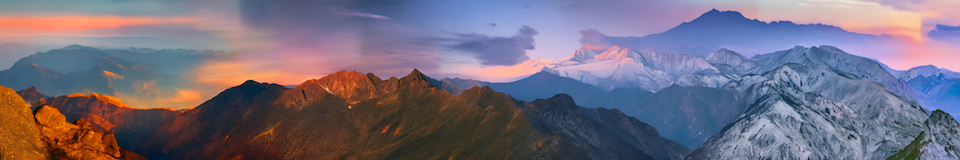}  \\
    \scriptsize{\makecell{Sync\\Diffusion}} & 
    \includegraphics[width=0.9\textwidth]{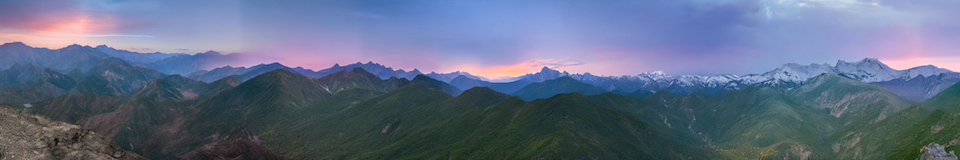} \\
    \bottomrule

    \multicolumn{2}{c}{\bigskip} \\

    \toprule
    \multicolumn{2}{c}{\textit{``Cartoon panorama of spring summer beautiful nature''}} \\
    \midrule

    \scriptsize{BLD~\cite{avrahami2023blended_latent}} & 
    \includegraphics[width=0.9\textwidth]{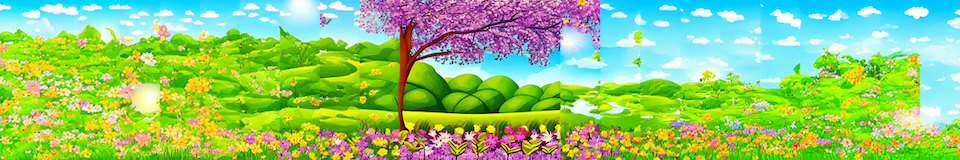} \\
    \scriptsize{MD~\cite{bartal2023multidiffusion}} & 
    \includegraphics[width=0.9\textwidth]{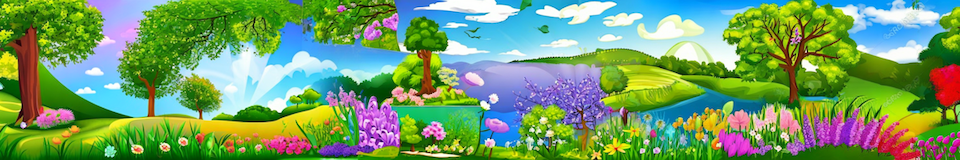}  \\
    \scriptsize{\makecell{Sync\\Diffusion}} & 
    \includegraphics[width=0.9\textwidth]{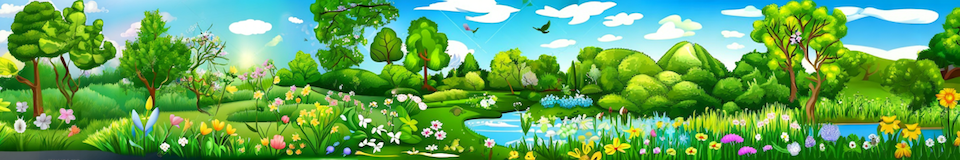} \\
    \bottomrule
\end{longtable}

%% file: neurips_2023.bbl
\begin{thebibliography}{10}

\bibitem{avrahami2023blended_latent}
Omri Avrahami, Ohad Fried, and Dani Lischinski.
\newblock Blended latent diffusion.
\newblock In {\em SIGGRAPH}, 2023.

\bibitem{avrahami2022blended}
Omri Avrahami, Dani Lischinski, and Ohad Fried.
\newblock Blended diffusion for text-driven editing of natural images.
\newblock In {\em CVPR}, 2022.

\bibitem{bartal2023multidiffusion}
Omer Bar-Tal, Lior Yariv, Yaron Lipman, and Tali Dekel.
\newblock {MultiDiffusion}: Fusing diffusion paths for controlled image generation.
\newblock In {\em ICML}, 2023.

\bibitem{kid2018}
Mikołaj Bińkowski, Danica~J. Sutherland, Michael Arbel, and Arthur Gretton.
\newblock Demystifying {MMD} {GAN}s.
\newblock In {\em ICLR}, 2018.

\bibitem{blattmann2023videoldm}
Andreas Blattmann, Robin Rombach, Huan Ling, Tim Dockhorn, Seung~Wook Kim, Sanja Fidler, and Karsten Kreis.
\newblock {Align your Latents}: High-resolution video synthesis with latent diffusion models.
\newblock In {\em CVPR}, 2023.

\bibitem{brock2019large}
Andrew Brock, Jeff Donahue, and Karen Simonyan.
\newblock Large scale {GAN} training for high fidelity natural image synthesis.
\newblock In {\em ICLR}, 2019.

\bibitem{brooks2022instructpix2pix}
Tim Brooks, Aleksander Holynski, and Alexei~A Efros.
\newblock Instructpix2pix: Learning to follow image editing instructions.
\newblock In {\em CVPR}, 2023.

\bibitem{chen2023text2tex}
Dave~Zhenyu Chen, Yawar Siddiqui, Hsin-Ying Lee, Sergey Tulyakov, and Matthias Nießner.
\newblock {Text2Tex}: Text-driven texture synthesis via diffusion models.
\newblock {\em arXiv preprint arXiv:2303.11396}, 2023.

\bibitem{chung2023dps}
Hyungjin Chung, Jeongsol Kim, Michael~Thompson Mccann, Marc~Louis Klasky, and Jong~Chul Ye.
\newblock Diffusion posterior sampling for general noisy inverse problems.
\newblock In {\em ICLR}, 2023.

\bibitem{chung2022mcg}
Hyungjin Chung, Byeongsu Sim, and Jong~Chul Ye.
\newblock Improving diffusion models for inverse problems using manifold constraints.
\newblock In {\em NeurIPS}, 2022.

\bibitem{dhariwal2021diffusion}
Prafulla Dhariwal and Alex Nichol.
\newblock Diffusion models beat {GAN}s on image synthesis.
\newblock In {\em NeurIPS}, 2021.

\bibitem{gal2023image}
Rinon Gal, Yuval Alaluf, Yuval Atzmon, Or~Patashnik, Amit~H. Bermano, Gal Chechik, and Daniel Cohen-Or.
\newblock An image is worth one word: Personalizing text-to-image generation using textual inversion.
\newblock In {\em ICLR}, 2023.

\bibitem{gao2023implicit}
Sicheng Gao, Xuhui Liu, Bohan Zeng, Sheng Xu, Yanjing Li, Xiaoyan Luo, Jianzhuang Liu, Xiantong Zhen, and Baochang Zhang.
\newblock Implicit diffusion models for continuous super-resolution.
\newblock In {\em CVPR}, 2023.

\bibitem{gatys2016styleloss}
Leon~A Gatys, Alexander~S Ecker, and Matthias Bethge.
\newblock Image style transfer using convolutional neural networks.
\newblock In {\em CVPR}, 2016.

\bibitem{ghosal2023texttoaudio}
Deepanway Ghosal, Navonil Majumder, Ambuj Mehrish, and Soujanya Poria.
\newblock Text-to-audio generation using instruction-tuned llm and latent diffusion model.
\newblock {\em arXiv preprint arXiv:2304.13731}, 2023.

\bibitem{goodfellow2014generative}
Ian Goodfellow, Jean Pouget-Abadie, Mehdi Mirza, Bing Xu, David Warde-Farley, Sherjil Ozair, Aaron Courville, and Yoshua Bengio.
\newblock Generative adversarial nets.
\newblock In {\em NeurIPS}, 2014.

\bibitem{gu2020giqa}
Shuyang Gu, Jianmin Bao, Dong Chen, and Fang Wen.
\newblock {GIQA}: Generated image quality assessment.
\newblock In {\em ECCV}, 2020.

\bibitem{hertz2022prompt}
Amir Hertz, Ron Mokady, Jay Tenenbaum, Kfir Aberman, Yael Pritch, and Daniel Cohen-Or.
\newblock Prompt-to-prompt image editing with cross attention control.
\newblock In {\em ICLR}, 2023.

\bibitem{hessel2021clipscore}
Jack Hessel, Ari Holtzman, Maxwell Forbes, Ronan~Le Bras, and Yejin Choi.
\newblock {CLIPScore:} a reference-free evaluation metric for image captioning.
\newblock In {\em EMNLP}, 2021.

\bibitem{heusel2018fid}
Martin Heusel, Hubert Ramsauer, Thomas Unterthiner, Bernhard Nessler, and Sepp Hochreiter.
\newblock {GAN}s trained by a two time-scale update rule converge to a local nash equilibrium.
\newblock In {\em NIPS}, 2018.

\bibitem{ho2020denoising}
Jonathan Ho, Ajay Jain, and Pieter Abbeel.
\newblock Denoising diffusion probabilistic models.
\newblock In {\em NeurIPS}, 2020.

\bibitem{ho2022video}
Jonathan Ho, Tim Salimans, Alexey Gritsenko, William Chan, Mohammad Norouzi, and David~J Fleet.
\newblock Video diffusion models.
\newblock In {\em ICLR}, 2022.

\bibitem{huang2023makeanaudio}
Rongjie Huang, Jiawei Huang, Dongchao Yang, Yi~Ren, Luping Liu, Mingze Li, Zhenhui Ye, Jinglin Liu, Xiang Yin, and Zhou Zhao.
\newblock {Make-An-Audio}: Text-to-audio generation with prompt-enhanced diffusion models.
\newblock {\em arXiv preprint arXiv:2301.12661}, 2023.

\bibitem{karras2020analyzing}
Tero Karras, Samuli Laine, Miika Aittala, Janne Hellsten, Jaakko Lehtinen, and Timo Aila.
\newblock Analyzing and improving the image quality of {StyleGAN}.
\newblock In {\em CVPR}, 2020.

\bibitem{kumari2023multiconcept}
Nupur Kumari, Bingliang Zhang, Richard Zhang, Eli Shechtman, and Jun-Yan Zhu.
\newblock Multi-concept customization of text-to-image diffusion.
\newblock In {\em CVPR}, 2023.

\bibitem{kwon2022diffusion}
Gihyun Kwon and Jong~Chul Ye.
\newblock Diffusion-based image translation using disentangled style and content representation.
\newblock In {\em ICLR}, 2023.

\bibitem{li2021srdiff}
Haoying Li, Yifan Yang, Meng Chang, Huajun Feng, Zhihai Xu, Qi~Li, and Yueting Chen.
\newblock Srdiff: Single image super-resolution with diffusion probabilistic models.
\newblock {\em Neurocomputing}, 2022.

\bibitem{lin2023magic3d}
Chen-Hsuan Lin, Jun Gao, Luming Tang, Towaki Takikawa, Xiaohui Zeng, Xun Huang, Karsten Kreis, Sanja Fidler, Ming-Yu Liu, and Tsung-Yi Lin.
\newblock Magic{3D}: High-resolution text-to-{3D} content creation.
\newblock In {\em CVPR}, 2023.

\bibitem{liu2023audioldm}
Haohe Liu, Zehua Chen, Yi~Yuan, Xinhao Mei, Xubo Liu, Danilo Mandic, Wenwu Wang, and Mark~D. Plumbley.
\newblock {AudioLDM}: Text-to-audio generation with latent diffusion models.
\newblock {\em arXiv preprint arXiv:2301.12503}, 2023.

\bibitem{lugmayr2022repaint}
Andreas Lugmayr, Martin Danelljan, Andres Romero, Fisher Yu, Radu Timofte, and Luc Van~Gool.
\newblock {Repaint}: Inpainting using denoising diffusion probabilistic models.
\newblock In {\em CVPR}, 2022.

\bibitem{meng2022sdedit}
Chenlin Meng, Yutong He, Yang Song, Jiaming Song, Jiajun Wu, Jun-Yan Zhu, and Stefano Ermon.
\newblock {SDE}dit: Guided image synthesis and editing with stochastic differential equations.
\newblock In {\em ICLR}, 2022.

\bibitem{mturk-2afc}
Rudimentary framework for running two-alternative forced choice (2afc) perceptual studies on mechanical turk.
\newblock \url{https://github.com/brownvc/mturk-2afc}.

\bibitem{nichol2021improved}
Alex Nichol and Prafulla Dhariwal.
\newblock Improved denoising diffusion probabilistic models.
\newblock In {\em ICML}, 2021.

\bibitem{nichol2021glide}
Alex Nichol, Prafulla Dhariwal, Aditya Ramesh, Pranav Shyam, Pamela Mishkin, Bob McGrew, Ilya Sutskever, and Mark Chen.
\newblock {GLIDE}: Towards photorealistic image generation and editing with text-guided diffusion models.
\newblock In {\em ICML}, 2022.

\bibitem{poole2022dreamfusion}
Ben Poole, Ajay Jain, Jonathan~T. Barron, and Ben Mildenhall.
\newblock {DreamFusion}: Text-to-{3D} using {2D} diffusion.
\newblock In {\em ICLR}, 2023.

\bibitem{ramesh2022dalle2}
Aditya Ramesh, Prafulla Dhariwal, Alex Nichol, Casey Chu, and Mark Chen.
\newblock Hierarchical text-conditional image generation with clip latents.
\newblock {\em arXiv preprint arXiv:2204.06125}, 2022.

\bibitem{ramesh2021dalle}
Aditya Ramesh, Mikhail Pavlov, Gabriel Goh, Scott Gray, Chelsea Voss, Alec Radford, Mark Chen, and Ilya Sutskever.
\newblock Zero-shot text-to-image generation.
\newblock In {\em ICML}, 2021.

\bibitem{richardson2023texture}
Elad Richardson, Gal Metzer, Yuval Alaluf, Raja Giryes, and Daniel Cohen-Or.
\newblock {TEXTure}: Text-guided texturing of {3D} shapes.
\newblock {\em arXiv preprint arXiv:2302.01721}, 2023.

\bibitem{rombach2022highresolution}
Robin Rombach, Andreas Blattmann, Dominik Lorenz, Patrick Esser, and Björn Ommer.
\newblock High-resolution image synthesis with latent diffusion models.
\newblock In {\em CVPR}, 2022.

\bibitem{ruiz2023dreambooth}
Nataniel Ruiz, Yuanzhen Li, Varun Jampani, Yael Pritch, Michael Rubinstein, and Kfir Aberman.
\newblock {DreamBooth}: Fine tuning text-to-image diffusion models for subject-driven generation.
\newblock In {\em CVPR}, 2023.

\bibitem{saharia2022palette}
Chitwan Saharia, William Chan, Huiwen Chang, Chris Lee, Jonathan Ho, Tim Salimans, David Fleet, and Mohammad Norouzi.
\newblock {Palette}: Image-to-image diffusion models.
\newblock In {\em SIGGRAPH}, 2022.

\bibitem{saharia2022photorealistic}
Chitwan Saharia, William Chan, Saurabh Saxena, Lala Li, Jay Whang, Emily Denton, Seyed Kamyar~Seyed Ghasemipour, Burcu~Karagol Ayan, S.~Sara Mahdavi, Rapha~Gontijo Lopes, Tim Salimans, Jonathan Ho, David~J Fleet, and Mohammad Norouzi.
\newblock Photorealistic text-to-image diffusion models with deep language understanding.
\newblock In {\em NeurIPS}, 2022.

\bibitem{saharia2021image}
Chitwan Saharia, Jonathan Ho, William Chan, Tim Salimans, David~J Fleet, and Mohammad Norouzi.
\newblock Image super-resolution via iterative refinement.
\newblock {\em IEEE Transactions on Pattern Analysis and Machine Intelligence}, 2022.

\bibitem{sohldickstein2015deep}
Jascha Sohl-Dickstein, Eric~A. Weiss, Niru Maheswaranathan, and Surya Ganguli.
\newblock Deep unsupervised learning using nonequilibrium thermodynamics.
\newblock In {\em ICML}, 2015.

\bibitem{song2022denoising}
Jiaming Song, Chenlin Meng, and Stefano Ermon.
\newblock Denoising diffusion implicit models.
\newblock In {\em ICLR}, 2021.

\bibitem{tang2023makeit3d}
Junshu Tang, Tengfei Wang, Bo~Zhang, Ting Zhang, Ran Yi, Lizhuang Ma, and Dong Chen.
\newblock {Make-It-{3D}}: High-fidelity {3D} creation from a single image with diffusion prior.
\newblock {\em arXiv preprint arXiv:2303.14184}, 2023.

\bibitem{tang2023mvdiffusion}
Shitao Tang, Fuyang Zhang, Jiacheng Chen, Peng Wang, and Yasutaka Furukawa.
\newblock {MVDiffusion}: Enabling holistic multi-view image generation with correspondence-aware diffusion.
\newblock {\em arXiv preprint arXiv:2307.01097}, 2023.

\bibitem{whang2022deblurring}
Jay Whang, Mauricio Delbracio, Hossein Talebi, Chitwan Saharia, Alexandros~G. Dimakis, and Peyman Milanfar.
\newblock Deblurring via stochastic refinement.
\newblock In {\em CVPR}, 2022.

\bibitem{xu2023dream3d}
Jiale Xu, Xintao Wang, Weihao Cheng, Yan-Pei Cao, Ying Shan, Xiaohu Qie, and Shenghua Gao.
\newblock {Dream3D}: Zero-shot text-to-{3D} synthesis using {3D} shape prior and text-to-image diffusion models.
\newblock In {\em CVPR}, 2023.

\bibitem{yang2022diffsound}
Dongchao Yang, Jianwei Yu, Helin Wang, Wen Wang, Chao Weng, Yuexian Zou, and Dong Yu.
\newblock {Diffsound}: Discrete diffusion model for text-to-sound generation.
\newblock {\em IEEE Transactions on Audio, Speech and Language Processing}, 2022.

\bibitem{zhang2023adding}
Lvmin Zhang and Maneesh Agrawala.
\newblock Adding conditional control to text-to-image diffusion models.
\newblock {\em arXiv preprint arXiv:2302.05543}, 2023.

\bibitem{zhang2023diffcollage}
Qinsheng Zhang, Jiaming Song, Xun Huang, Yongxin Chen, and Ming-Yu Liu.
\newblock {DiffCollage}: Parallel generation of large content with diffusion models.
\newblock In {\em CVPR}, 2023.

\bibitem{zhang2018lpips}
Richard Zhang, Phillip Isola, Alexei~A. Efros, Eli Shechtman, and Oliver Wang.
\newblock The unreasonable effectiveness of deep features as a perceptual metric.
\newblock In {\em CVPR}, 2018.

\end{thebibliography}
